\documentclass[10pt, letterpaper, onecolumn, draftclsnofoot]{IEEEtran}

\usepackage{multirow}
\usepackage{makecell} 
\usepackage{bm}

\usepackage{array}

\usepackage[]{todonotes}
\usepackage[utf8]{inputenc}
\usepackage[T1]{fontenc}
\usepackage{lmodern}

\usepackage{subcaption}  

\usepackage{graphicx}
\usepackage{xcolor}

\usepackage{gensymb}

\usepackage{hyperref}
\hypersetup{
	colorlinks=true,
	linkcolor=blue,
	filecolor=magenta,      
	urlcolor=cyan,
	pdftitle={An All-In-One, Low-Cost Photogrammetry Rig for 3D Plant Modelling and Phenotyping},
	pdfpagemode=FullScreen,
}

\usepackage{amsmath}
\usepackage{amssymb}
\usepackage{amsfonts}

\usepackage{algorithm, tabularx}
\usepackage{algpseudocode}

\algdef{SE}[SUBALG]{Indent}{EndIndent}{}{\algorithmicend\ }%
\algtext*{Indent}
\algtext*{EndIndent}

\algblock[Class]{Class}{End}

\title{A Low-Cost Photogrammetry System\\
	for 3D Plant Modeling and Phenotyping}
\author{
	\IEEEauthorblockN{Joe Hrzich\IEEEauthorrefmark{1}, Michael A. Beck\IEEEauthorrefmark{1}\textsuperscript{,}\IEEEauthorrefmark{2}, Christopher P. Bidinosti\IEEEauthorrefmark{1}\textsuperscript{,}\IEEEauthorrefmark{3},\\ Christopher J. Henry\IEEEauthorrefmark{4}, Kalhari Manawasinghe\IEEEauthorrefmark{5}, Karen Tanino\IEEEauthorrefmark{5}}\\
	
	\IEEEauthorblockN{\IEEEauthorrefmark{2}Corresponding author} \\
    \IEEEauthorblockA{\IEEEauthorrefmark{1}Department of Applied Computer Science, University of Winnipeg, Winnipeg, Canada \texttt{hrzich-j@webmail.uwinnipeg.ca}, \texttt{m.beck@uwinnipeg.ca}} \\
	\IEEEauthorblockA{\IEEEauthorrefmark{3}Department of Physics, University of Winnipeg, Winnipeg, Canada \newline
		\texttt{c.bidinosti@uwinnipeg.ca}} \\
	\IEEEauthorblockA{\IEEEauthorrefmark{4}Department of Computer Science, University of Manitoba, Winnipeg, Canada \texttt{christopher.henry@umanitoba.ca}} \\
	\IEEEauthorblockA{\IEEEauthorrefmark{5}Department of Plant Sciences, University of Saskatchewan, Saskatoon,  Canada \texttt{fhw048@usask.ca, karen.tanino@usask.ca}} \\
}

\date{\today}

\begin{document}
	
	\maketitle

	\begin{abstract}
		We present an open-source, low-cost photogrammetry system for 3D plant modeling and phenotyping. The system uses a structure-from-motion approach to reconstruct 3D representations of the plants via point clouds. Using wheat as an example, we demonstrate how various phenotypic traits can be computed easily from the point clouds. These include standard measurements such as plant height and radius, as well as features that would be more cumbersome to measure by hand, such as leaf angles and convex hull. We further demonstrate the utility of the system through the investigation of specific metrics that may yield objective classifications of erectophile versus planophile wheat canopy architectures.
	\end{abstract}

    
	\section{Introduction}\label{sec:introduction}
	The measurement of phenotypic data is a key component of plant research and is essential in breeding programs for the development of varieties that are, for example, more resilient to a changing climate and plant diseases. 
    High throughput phenotyping -- the extraction of plant features in a fast and automated way~\cite{castro2021,akthar2024} -- is expected to play an ever increasing role in evaluating plant varieties and assessing plant health. 
    However, cost and complexity remain barriers to entry for many researchers, resulting in the continued practice of manual measurement and visual inspection, which are time consuming and prone to error and bias.
    Fortunately, the price of embedded systems has steeply decreased in the last decades, enabling the creation of systems that fulfill complex tasks at low costs. 
    
    Following this trend and addressing the needs of the phenotyping community, we designed a low-cost plant photogrammetry system.  It is based on the principle of structure from motion (SfM), ultilizing a photography turntable and several cameras to obtain 3D models
    from collections of 2D images taken from different perspectives.
    It was built from commercially available off-the-shelf materials and all software is freely available, keeping the overall price below \$3,000 CAD.
    The system enables one to extract a multitude of plant metrics quickly and easily
    from 3D point cloud representations, which, compared to 2D imaging techniques, contains a richer set of features. To demonstrate the capabilities of the system we use it to extract a variety of phenotypic metrics of wheat plants. We further relate these metrics to numerical classifications of the wheat canopy architecture (i.e.\ erectophile versus planophile), a task  which is typically performed by 
    human visual inspection and the assignment of a value on a subjective scale.
	
	
	Overall the  main contributions of this work are the following: 
	\begin{itemize}
		\item Description and construction of a photogrammetry system that is open-source and low-cost. 
		\item Providing software for the system that allows push-button data generation.
		\item Development of algorithms for cleaning and pre-processing the obtained point clouds and to extract phenotypical information from point clouds.
		\item Demonstrating the system's capability by analyzing wheat plants of different varieties with a focus on traits that describe their canopy architecture. 
	\end{itemize}
	
	The  paper is organized as follows. Section \ref{sec:related_works} describes related work with respect to imaging systems and methods to obtain 3D data. Section \ref{sec:hardware} and Section \ref{sec:software} describe the hardware and software components of the system, respectively. The collection of a 3D wheat dataset and the traits extracted from these point clouds is described in Section \ref{sec:measurements}. In Section \ref{sec:data_analysis} we analyze these features and compare them to the manual labelling of the plants performed by a plant expert. The paper is concluded in Section \ref{sec:conclusion}.
	
	\section{Related Works}\label{sec:related_works}
	Compared to 2D imaging, 3D data offers a combined perspective of plant structures ranging from the surface to internal segments \cite{Vylder}. The ability to amalgamate data from various viewing angles offers insights that can be challenging to achieve with a 2D model alone. For example, 3D data has been used to address challenges in resolving occlusions \cite{Salter}, estimating biomass \cite{KEIGHTLEY2010305}, estimating yield \cite{Paulus2013}, identifying plant diseases \cite{Nagasubramanian2019}, quantification of architectural traits such as leaf area and canopy volume \cite{Salter, Paulus2014}, and measuring plant responses to stressors \cite{Khanna2019}.
	
	To obtain 3D data, several instruments and techniques are available. We first discuss some of the alternatives to photogrammetry systems and examples of their applications in phenotyping. LiDAR uses a pulsed laser to measure distances, providing 3D data that includes internal plant structures such as branching architecture and leaf density \cite{Paulus2019}. For example, the authors of Ref.~\cite{HOSOI2009151} used a high-resolution portable scanning LiDAR to estimate wheat vertical plant area density profiles at different growth stages. In another study, researchers developed a method for mapping the leaf area index (LAI) using ground-based laser rangefinders, a LiDAR-based method—mounted on a vehicle \cite{Gebbers}. However, the high costs associated with LiDAR can limit its application.
	
	Structured light approaches use information of how a known light pattern is distorted when projected onto an object to infer 3D information. This can be used for trait analysis such as determining the number of leaves, plant height, leaf size, and internode distances \cite{Nguyen}. The authors of Ref.~\cite{PaulusLT} utilized a high-precision laser scanning system to non-invasively capture the 3D architecture of barley plants, focusing on traits such as leaf area, stem height, and overall plant volume. Structured light approaches are typically limited by their shorter working distances \cite{gibbs2016approaches, Paulus2019}, can impact the natural physiological state of the plant, and alter its original texture or color \cite{Lu_3Dplant}.
	
	Stereo vision systems employ multiple cameras to capture different perspectives, providing surface information at a lower cost; however, they can struggle with overlapping structures. It is the most similar approach to SfM photogrammetry reconstructions, which can be seen as a generalization of stereo vision systems. Examples of the use of stereo vision systems for plant research include the reconstruction of sorghum plant architectures~\cite{BaoStereoVision2019}, the estimation of sorghum stem diameters~\cite{xiang2020phenostereo}, and grapevine growth~\cite{klodtFieldPhenotypingGrapevine2015a} . 
	
	Photogrammetry based on SfM is a cost-effective  way to obtain 3D data that utilizes relatively simple sensors and hardware. SfM is utilized in Ref.~\cite{Paproki2012} to analyze cotton plant stem heights, leaf widths, and leaf lengths. In Ref.~\cite{Moeckel2018}, unmanned aerial vehicles (UAV) are used to obtain 3D data from different vegetable crops. SfM is also used by the authors of Ref.~\cite{Shi2016} to measure maize height and canopy structures.  Further examples include the study of strawberries \cite{He}, peppers \cite{Heijden}, soybeans \cite{biskup2007stereo}, maize and sorghum \cite{Shafiekhani2017}, grapevines \cite{MILELLA2019293}, chickpeas \cite{Salter}, and weeds \cite{weedphotogrammetry2018}. Some of these works rely on manual imaging from different angles (i.e., performed by a person), while others use some degree of automation. The automated systems listed rely on hardware that is not low-cost, such as DSLR cameras and/or accentuated actuators. Often the systems are designed for a very specific application or crop in mind. 
    More closely related to the open-source, more general purpose system described here 
    are the following: Lu et al.~\cite{Lu_3Dplant} focus their study on determining the optimal arrangement of cameras to obtain the best possible reconstruction.  They employ 10 DSLR cameras, however, and  cost-efficiency is not a priority. The works of Doan and Nguyen \cite{DOAN2024337} and Plum and Labonte \cite{plumScAntOpensourcePlatform2021} both present low-cost photogrammetry systems, but the purpose is for imaging small objects such as insects. 
	
	\section{Description of Hardware}\label{sec:hardware}
\begin{figure}[t]
		\centering
		\subfloat{{\includegraphics[width=0.4\columnwidth]{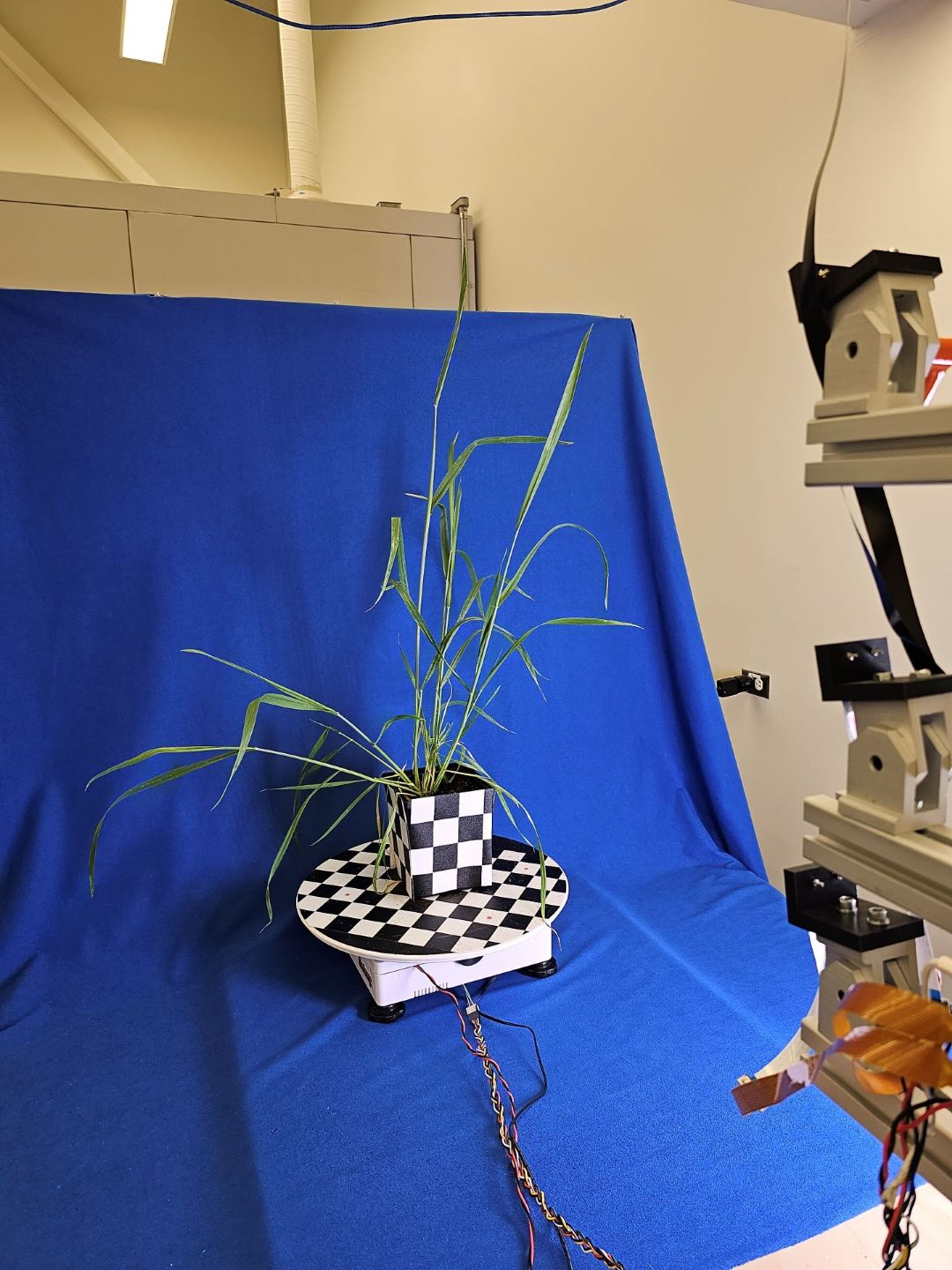} }}%
		\subfloat{{\includegraphics[width=0.4\columnwidth]{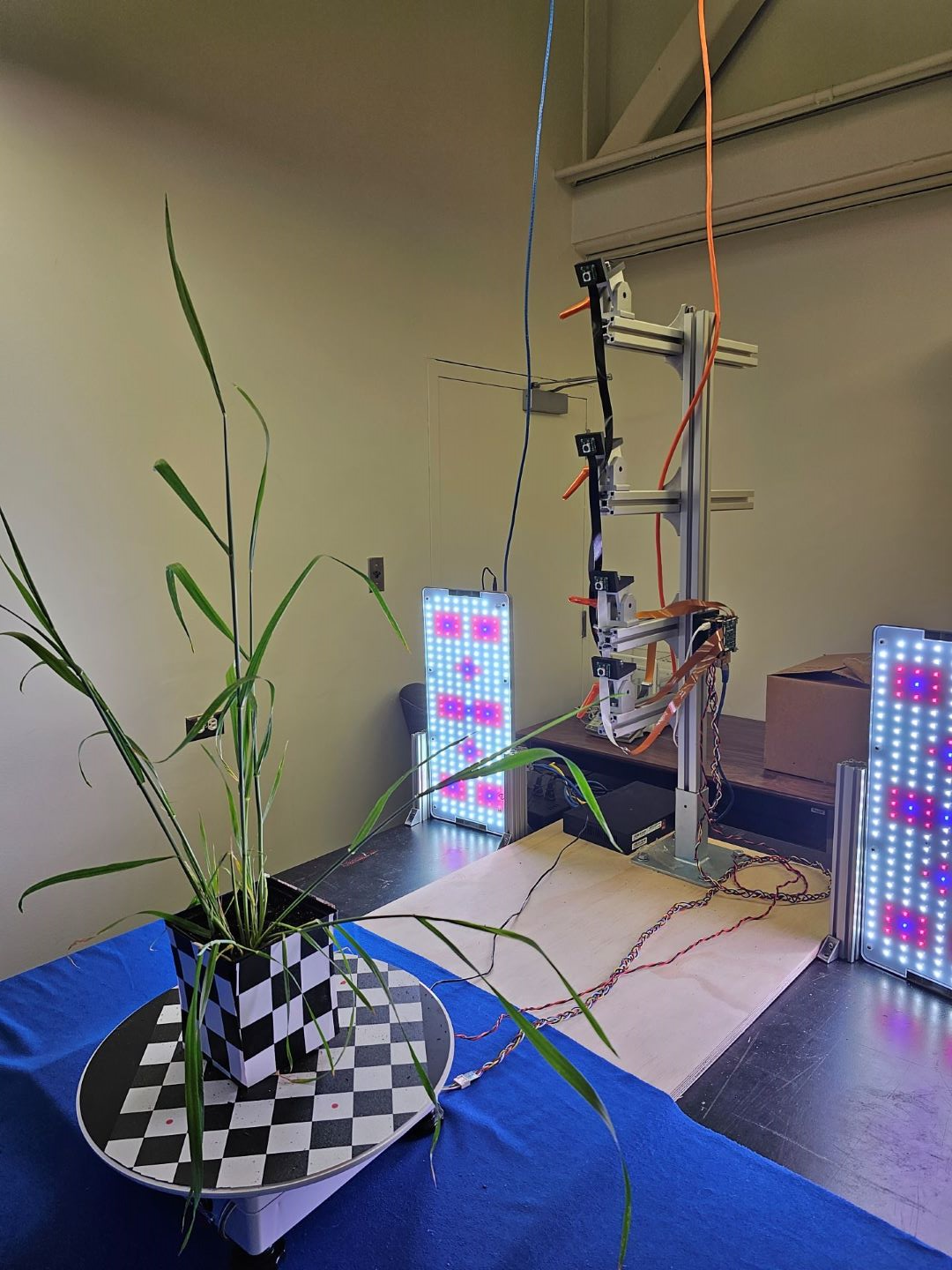} }}%
		\caption[Complete assembly of the photogrammetry rig]{
			(Left) A view of turntable with an example wheat plant and blue backdrop. In the foreground three of the four auto-focus cameras can be seen. (Right) The same system viewed from the side of the backdrop, showing the aluminum extrusion stand with the four cameras, the Raspberry Pi, and lights for additional illumination.}
		\label{fig:system-overview}%
	\end{figure}

	In this section we describe the hardware components of the photogrammetry system. 
    Figure~\ref{fig:system-overview} shows an overview of the system, specifically its structural framework, most of its electronic components, the arrangement of additional LED lighting, the matte blue backdrop, and the motorized turntable with an added checkerboard pattern. Table~\ref{tab:hardware-components} provides a breakdown of the hardware components along with pricing. The total cost of the system was just under \$3,000 Canadian.

	\begin{table*}
		\centering
		\begin{tabular}{p{0.1\linewidth}  p{0.08\linewidth}   >{\raggedright\arraybackslash}p{0.1\linewidth} p{0.05\linewidth}  >{\raggedright\arraybackslash}p{0.2\linewidth}  >{\raggedright\arraybackslash}p{0.25\linewidth}}
			\hline
			\textbf{Part} & \textbf{Brand}  & \textbf{Model} & \textbf{Price} & \textbf{Specs} & \textbf{Description} \\ 
			\hline
			Single board computer & Raspberry Pi & Raspberry Pi 4 Model B & \$249.99 & Quad-core Cortex-A72, 4GB RAM, 40 pin GPIO header, 2-lane MIPI CSI camera port & Compact, low-cost computer board ideal for embedded applications \\
			\hline
			4 RGB\newline Cameras & Arducam & 64MP Autofocus Quad-Camera Kit & \$199 & WiFi/Bluetooth, Res: 4056 x 3040 px, FOV diagonal: 84 $^{\circ}$ & High-resolution, auto-focus RGB cameras for detailed image capture \\
			\hline
			Featureless Background & Rose Brand & 62" Poly Pro & \$15.20 /yard & Fabric, Chroma Key Blue & Wrinkle-resistant fabric providing a uniform backdrop for imaging \\
			\hline
			Stepper\break Motor & Adafruit & Stepper Motor HAT - mini kit & \$27.95 & TB6612 chipset, 4.5VDC to 13.5VDC, unipolar or bipolar & Motor HAT (Hardware Attached on Top) designed for precise motion control with the Raspberry Pi \\
			\hline
			Turntable & Ortery & PhotoCapture 360 & \$1,200 & 11.4 and 15.7~in diameter platforms,  & Software-controlled rotary table for consistent image capture angles \\
			\hline
			Aluminum Extrusion & Misumi & & \$950 & aluminum alloy, Square shape, 40mm main frame and arms size, four side slots & Frame with stationary mounts, angle brackets, nuts and screws \\
			\hline
			All-Purpose LED Grow\newline Lights & Root Farm &  & \$145 each & Each 45~W, 25 cm (10 in) x 55 cm (21.5 in), adjustable legs with 360° swivel& Two growing light panels, consisting of 3-band LEDs, repurposed for additional photography lighting. \\
			\hline
		\end{tabular}
		\caption[Cost of hardware and parts for the photogrammetry rig.]{Components list for  the photogrammetry system.  (Prices  in Canadian dollars.)}
		\label{tab:hardware-components}
	\end{table*}

	At the core of the system is a Raspberry Pi 4 4GB single-board computer (RPi) \cite{RaspberryPi4ModelB4GB}, coupled with the 64MP Autofocus Synchronized Quad-Camera Kit \cite{64mp_autofocus_quad_camera_kit}. This kit allows us to attach and control 4 cameras with autofocus feature. The RPi controls a motorized turntable (Ortery PhotoCapture 360 \cite{OrteryPhotoCapture360M}) through an Adafruit DC and Stepper Motor HAT \cite{adafruit_motor_hat}. We use a 32GB microSD card for primary data storage on the RPi.
	
	The RPi was selected as the primary control and data acquisition system due to its cost-effectiveness, user-friendliness, wide community support, and expandability. By default, the RPi interfaces with a single camera only, thus we use the Arducam 64MP Autofocus Quad-Camera Kit HAT to extend the system to four cameras instead. The kit includes four 64MP cameras, which automatically adjust focus across various distances. Each camera module is mounted on adjustable brackets, allowing them to be easily pointed at the imaging target.
	
\begin{figure}
		\centering
		\includegraphics[width=1.0\linewidth]{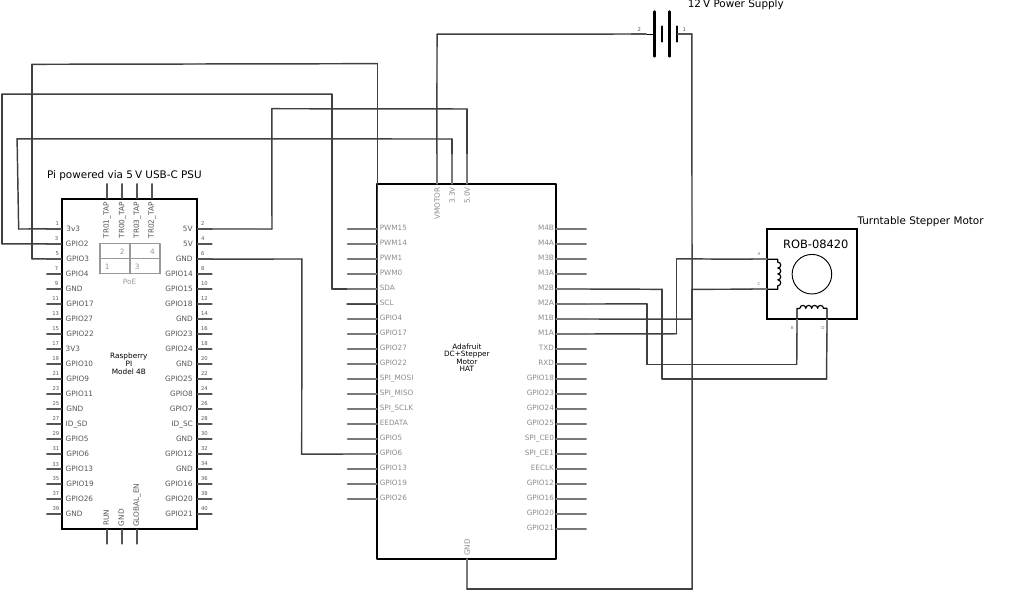}
		\caption{Schematic of the modified wiring required to drive the the turntable stepper motor via the RPi and Adafruit DC \& Stepper Motor HAT.}
		\label{fig:wiring_diagram}
	\end{figure}

	The system uses the commercial PhotoCapture 360 turntable~\cite{OrteryPhotoCapture360M}. It has a circular platform of 15.8 in (39.8 cm) diameter and the motorized rotation has a precision of ±1 degree. An additional feature of the turntable is that it supports weights up to 11 kg (more than needed here), while maintaining uniform and stable rotation. The turntable has peripheral rollers to evenly distribute the load, mitigating oscillations, which is beneficial, as plants are very susceptible to movements. 
    We added a black and white checkered pattern to the surface of the turntable (see Fig.~\ref{fig:system-overview}) that enhances visual alignment between individual images and thus is beneficial for the 3D reconstruction using SfM.
    Driving the turntable motor directly from an RPi is not possible, due to power constraints on the RPi; thus, to control the rotary table, we reconfigured its internal wiring to connect with an Adafruit DC \& Stepper Motor HAT~\cite{adafruit_motor_hat}, as shown in Fig.~\ref{fig:wiring_diagram}. 
    This board is accessed by the RPi over the Inter-Integrated Circuit (I2C) serial communication bus and uses pulse-width modulation to control the rotary table's motor. 
    Finally, we note  that the turntable is the most expensive piece of hardware in our system (see Table~\ref{tab:hardware-components}), and it is very likely that a rotating platform with lesser specifications is also sufficient for accurate 3D reconstruction of the plants.

	The frame of the imaging system that holds the four cameras and the RPi is constructed from standard T-slot aluminum extrusion. This choice was made for its modularity: the camera mount positions and angles are adaptable by T-slot brackets and fasteners. The main structural frame is 1 meter in length and has four arms (25-50 cm in length) each dedicated to holding one camera. This arrangement allows  the camera to be positioned in an arc, which together with a full rotation of the rotary table, capture a hemispherical view of the plant. The tilt of the cameras is easily adjustable by levers attached to their mounting brackets. The T-slot design also allows straightforward modifications, such as adding cameras or adjusting mounts to suit different sized plants, thereby enhancing the versatility of the system. Adapter pieces, which connect the RPi and camera modules with the aluminum extrusion, were 3D-printed. 

	The background in a photogrammetry setup is critical to the success of the imaging process. Patterns or distinct colors in the background can lead to misinterpretations as they do not rotate in alignment with the object being imaged. Thus, we use a featureless, blue matte, wrinkle-resistant backdrop, ensuring that the reconstruction algorithm is focused on the subject. In addition to the room lighting, two additional LED arrays were employed to further illuminate the plants without casting unwanted shadows or highlights on the backdrop. 
	
	\section{Software architecture}\label{sec:software}
	The software of the photogrammetry system is divided into four modules: 1) hardware control with a graphical user interface (GUI); 2) automated image acquisition and masking; 3) construction of 3D point clouds; 4) processing of point clouds. We present here a high-level overview of each of these modules. A full implementation is available on GitHub \cite{githubrepo}.

	\subsection{Hardware control}
\begin{figure*}[h]
		\centering
		\includegraphics[width=0.6\linewidth]{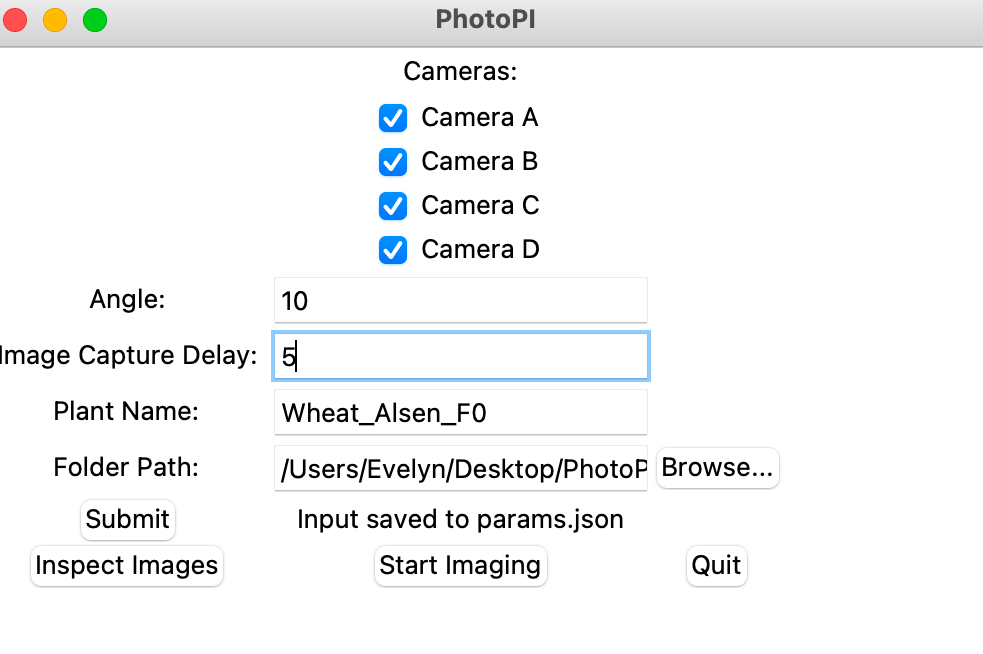}
		\caption{Screenshot of the system's graphical user interface, showing the settings the user can adjust, including which cameras to use, how far the turntable rotates between images taken and the wait time after rotation for any movements of the plant to settle.}
		\label{fig:GUI-screenshot}
	\end{figure*}

	The control software, written in Python, executes on the client side, i.e., a PC connected to the RPi. A GUI, based on the Tkinter library, is used to configure settings, initiate the image capture process, and inspect images. For this the software establishes SSH (Secure Shell) and SFTP (SSH File Transfer Protocol) connections with the RPi, using the Python package Paramiko \cite{paramiko}. Figure \ref{fig:GUI-screenshot} shows a screenshot of the GUI application, highlighting camera options, turntable adjustments, plant labeling, reviewing captured images and initiating the imaging sequence.
	
	The script begins by launching the \textit{InputGUI} class, which sets up the main Tkinter window for the PhotoPI application. The application discovers the Pi's hostname (\textit{raspberrypi.local}) and resolves it to its corresponding local IP address. Once the hostname and credentials are successfully obtained, the \textit{NetworkManager} class facilitates the SSH and SFTP connections, enabling the GUI to execute remote commands and transfer configuration files. For efficient image management, the script creates and manages specific directories—such as \textit{mask\_folder}, \textit{images\_folder}, and \textit{inspect\_folder} which are designated for storing various types of images. This approach ensures communication between the user interface and the RPi, allowing functionalities like image capture, stepper motor control, and system operations to be performed reliably and securely.
	
	After the initial setup, the GUI class is initiated, setting up the user interface framework. It includes functions such as \textit{browse\_folder()}, \textit{submit()}, \textit{inspect()}, and \textit{start\_imaging()}. Central to the script’s functionality are the inspection and image capture features, which handle capturing images using the connected cameras. The script leverages \textit{libcamera}, an open-source camera stack and framework for Linux, to capture images on the RPi. Additionally, it utilizes \textit{i2cset} to switch between cameras. 
	
	Following the image capture process, the script translates the specified angle parameter into stepper motor steps, directing the RPi to rotate the turntable accordingly. Finally, the script transfers all captured images to the local machine for inspection and storage using SFTP. The GUI interacts with a JSON configuration file to read and write settings, including hostnames, IP addresses, camera configurations, turntable settings, and folder paths. Table \ref{tab:config-parameters} provides an overview of the parameters for the control script, highlighting the functionalities of each parameter within the system.

	\begin{table}[h]
		\centering
        \normalsize
		\begin{tabular}{p{0.45\linewidth}p{0.45\linewidth}}
			\hline
			\textbf{Parameter} & \textbf{Description} \\
			\hline
			camera\_a, camera\_b, camera\_c, camera\_d & Boolean variables to enable/disable\newline corresponding cameras \\
			\hline
			angle & Rotation angle of the turntable in degrees \\
			\hline
			seconds & Time delay before taking another set of images \\
			\hline
			plant\_name & Name of the plant being imaged \\
			\hline
			folder\_path & Path to the folder for saving images \\
			\hline
			main\_hostname, main\_ip\_address,\newline pi\_hostname, pi\_ip\_address & Network configurations for SSH and SFTP connections \\
			\hline
			mask\_folder, images\_file, inspect\_folder & Directories for storing different types of images \\
			\hline
			SSH and SFTP Clients & Components for remote connections and file transfers \\
			\hline
			User Interface Functions & Functions for GUI interactions like folder\newline browsing, form submission, inspect images,\newline image capture \\
			\hline
			
		\end{tabular}
		\caption[Photogrammetry Rig Control Script Parameters]{Parameters of the control script and their respective functionalities.}
		\label{tab:config-parameters}
	\end{table}
	
\begin{figure*}
		\centering
		\includegraphics[width=0.9\linewidth]{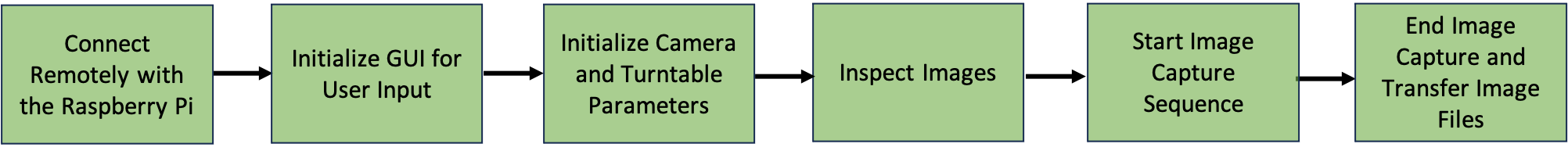}
		\caption{Flowchart of individual steps involved in collecting images with the photogrammetry system.}
		\label{fig:software-overview}
	\end{figure*}

	 Figure \ref{fig:software-overview} illustrates the entire pipeline for image capture using the PhotoPI control script. After positioning the plant of interest on the turntable, the user establishes a connection between their PC and the RPi and starts the GUI application. The user can then retrieve test images to ensure that the cameras are correctly positioned, providing a clear view of the plant. This can be done by capturing individual images or in bulk after the turntable completes each rotation. Subsequently, the user initiates the image capture sequence. Upon completing a full rotation, the system automatically stops the turntable and cameras. The user then has the option to review the captured images through the GUI. If necessary, the user can repeat the image capture process after making any required adjustments. Once the image quality meets the desired standards, the data transfer process can commence, automatically transferring the images from the RPi to the connected computer or a remote server for storage.

	\subsection{Image acquisition}
	
	During image capture, the system creates three distinct folders for data handling. The first folder stores the raw images as the cameras capture them. The second, labeled \textit{mask images}, contains processed images used for segmentation to separate the subject from its background. The third folder, labeled \textit{inspect images}, contains images used for quality and completeness checks. These images are inspected by the user to ensure the subject is positioned correctly. This allows for the early detection and correction of issues like blurring or improper exposure.
	
	The system employs a naming convention for the captured images in the form of 
    \[\text{PlantType\_CameraID\_CaptureDate\_SequenceNumber.}\] 
    This structured format includes the type of plant being imaged, the specific camera used for the capture (e.g., Camera A, B, C, D), the date of capture in a \textit{YYYMMDD} format, and a sequential number indicating the order in which the image was taken. For example, a file named \textit{Wheat\_CameraB\_20230718\_001} would represent the first image of a wheat plant, captured by Camera B on July 18, 2023. 
	
\begin{figure*}[t]
		\centering
		\subfloat[\centering Camera A (Raw)]{
			\includegraphics[width=0.23\textwidth]{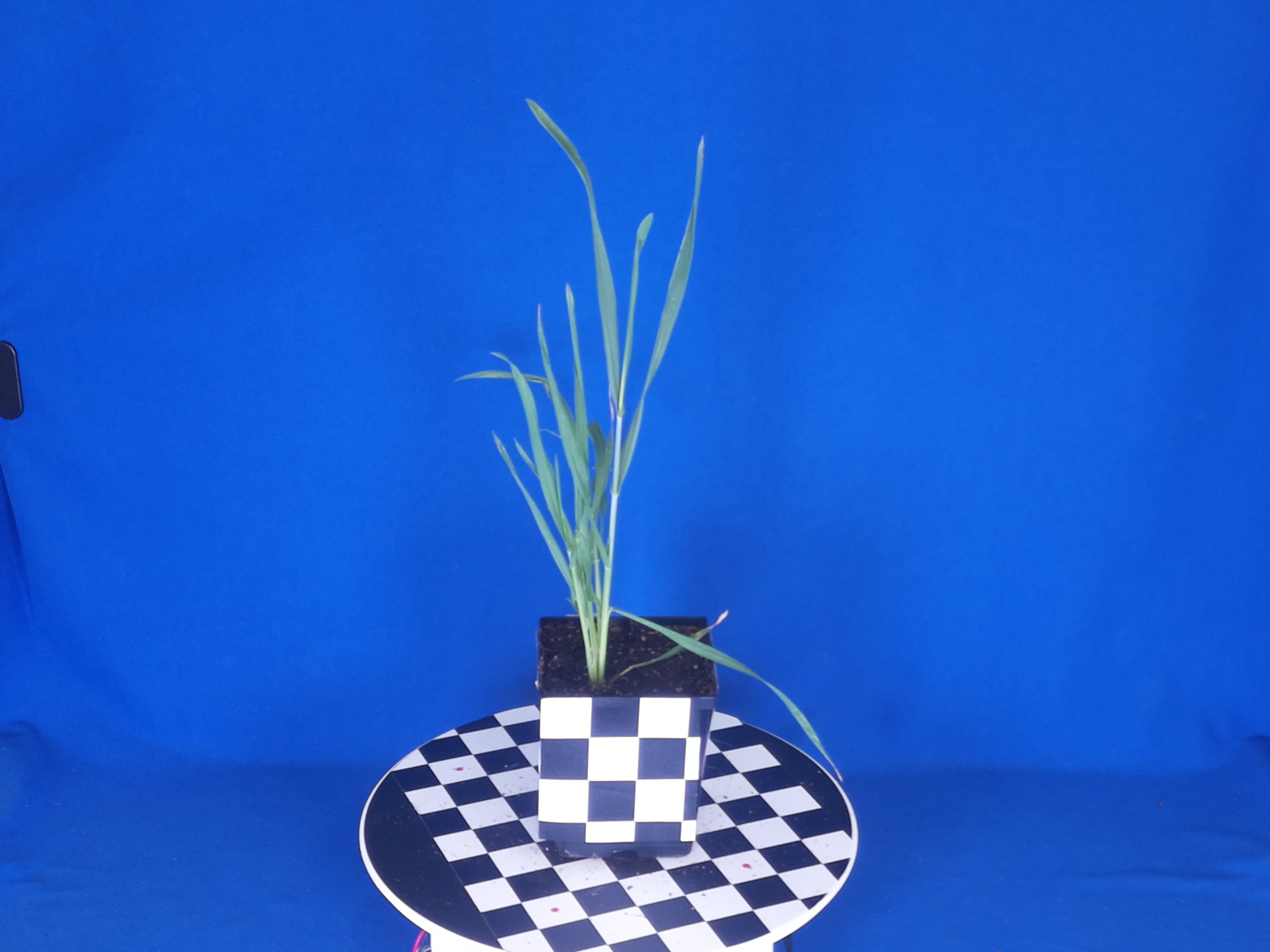}
		}
		\subfloat[\centering Camera B (Raw)]{
			\includegraphics[width=0.23\textwidth]{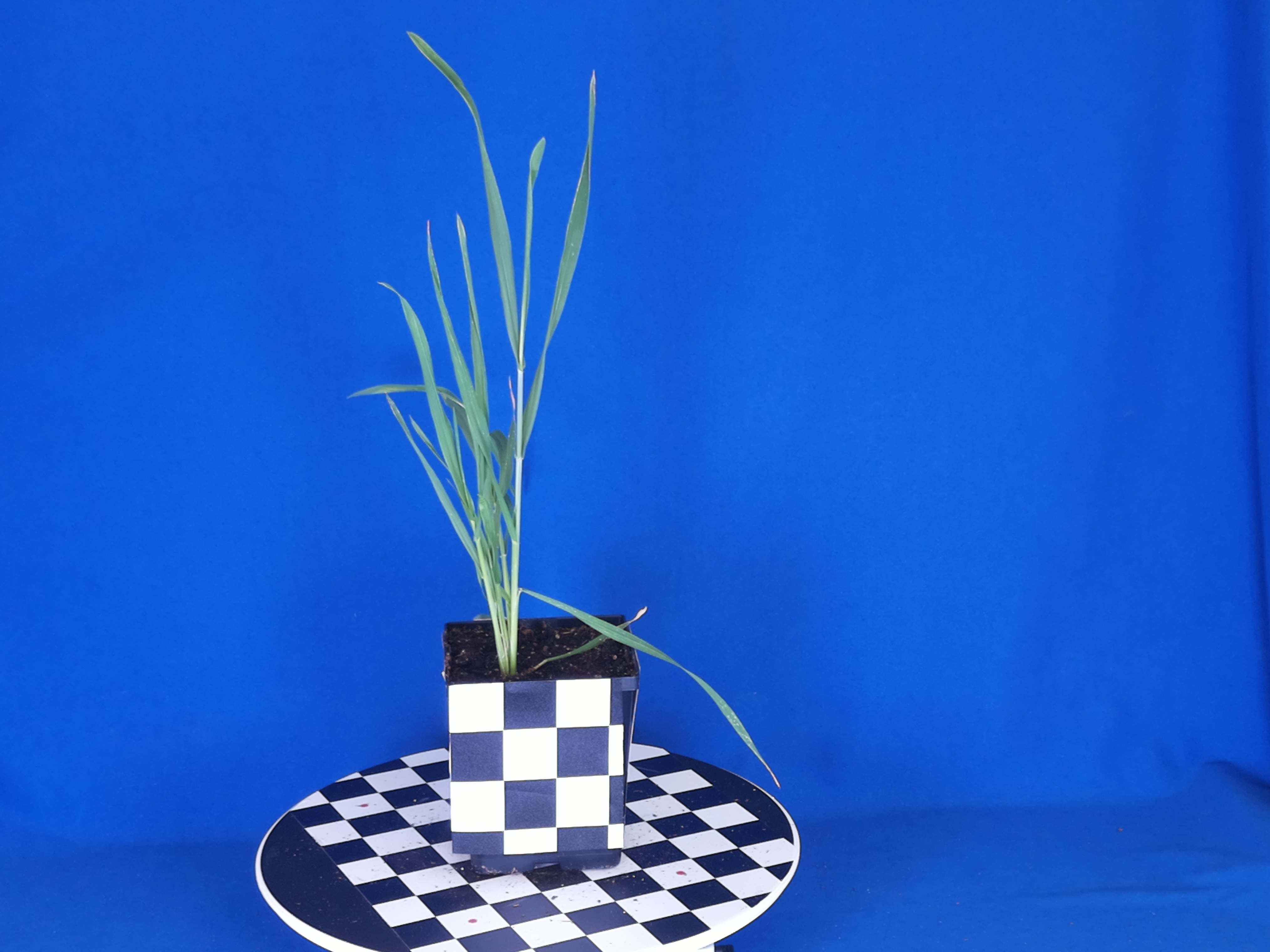}
		}
		\subfloat[\centering Camera C (Raw)]{
			\includegraphics[width=0.23\textwidth]{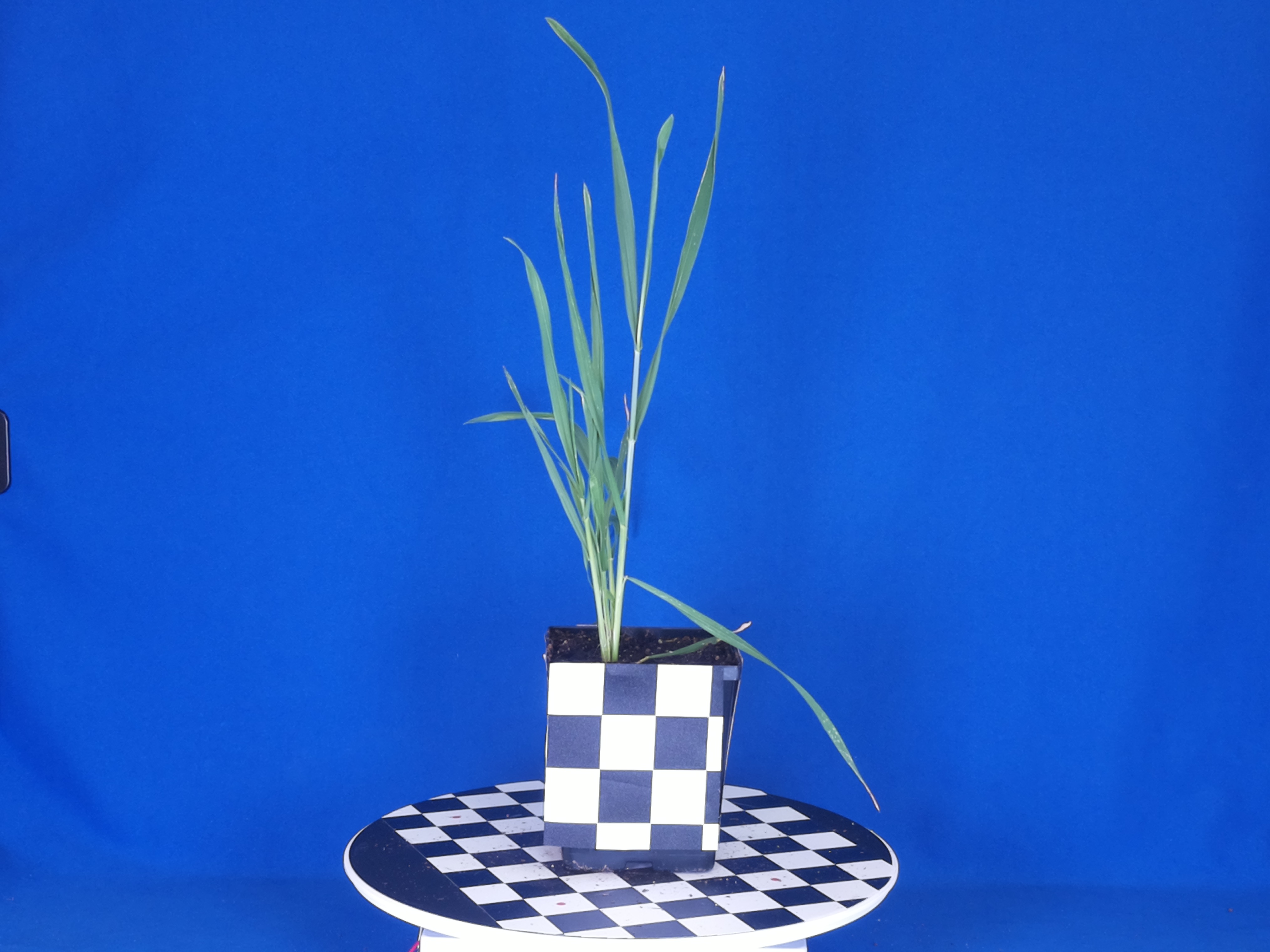}
		}
		\subfloat[\centering Camera D (Raw)]{
			\includegraphics[width=0.23\textwidth]{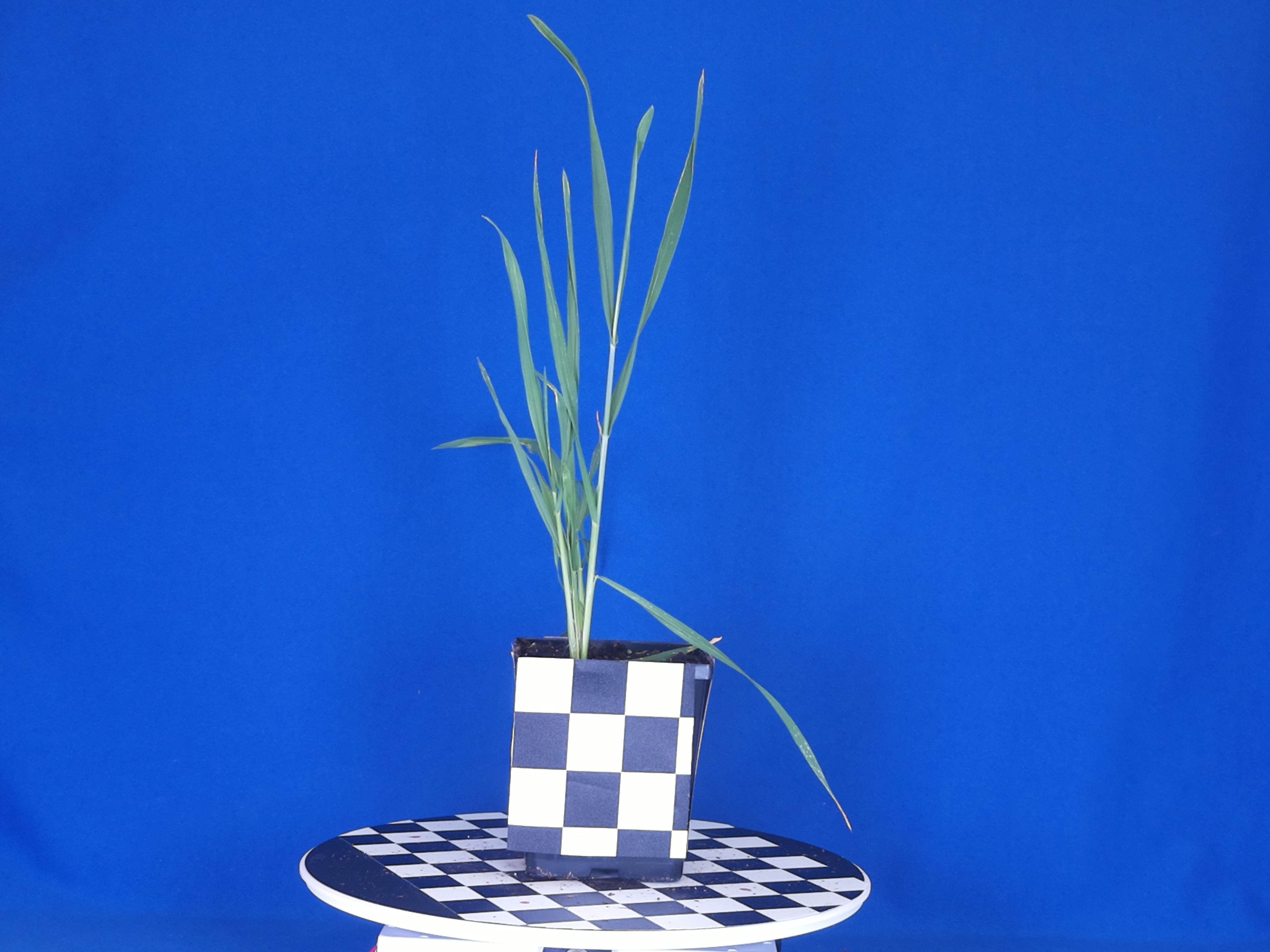}
		}
		\\ 
		\subfloat[\centering Camera A (Binary)]{
			\includegraphics[width=0.23\textwidth]{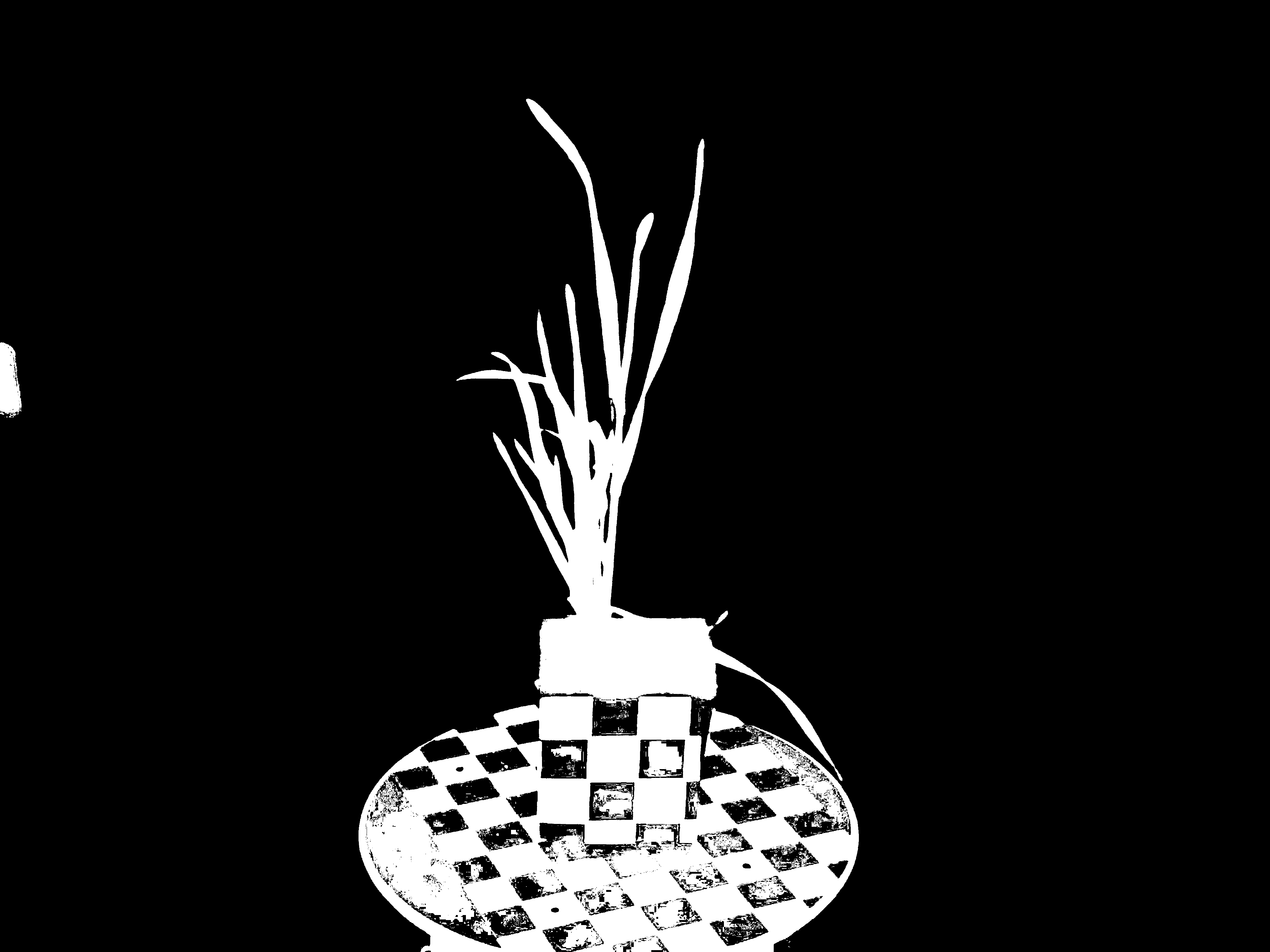}
		}
		\subfloat[\centering Camera B (Binary)]{
			\includegraphics[width=0.23\textwidth]{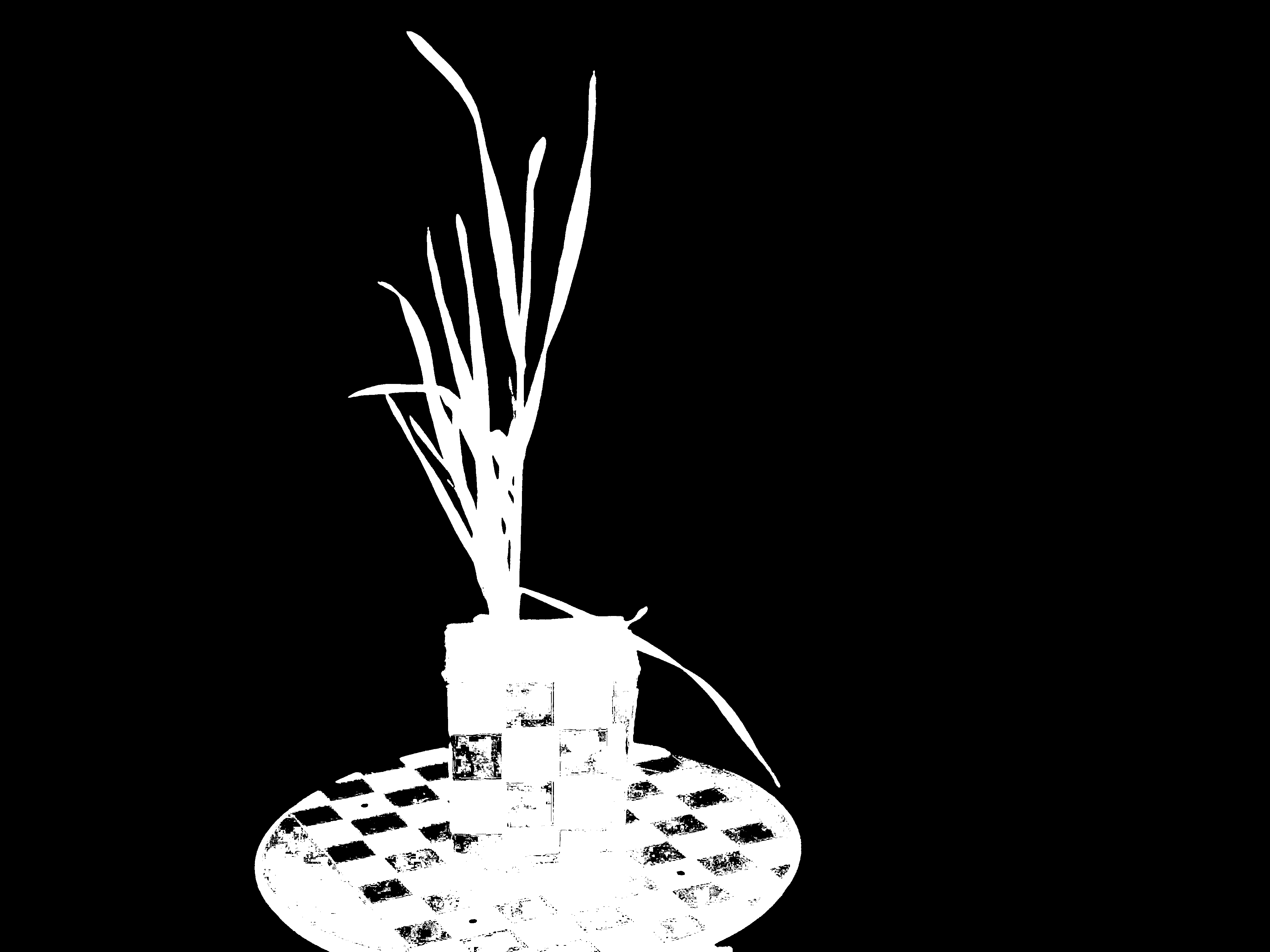}
		}
		\subfloat[\centering Camera C (Binary)]{
			\includegraphics[width=0.23\textwidth]{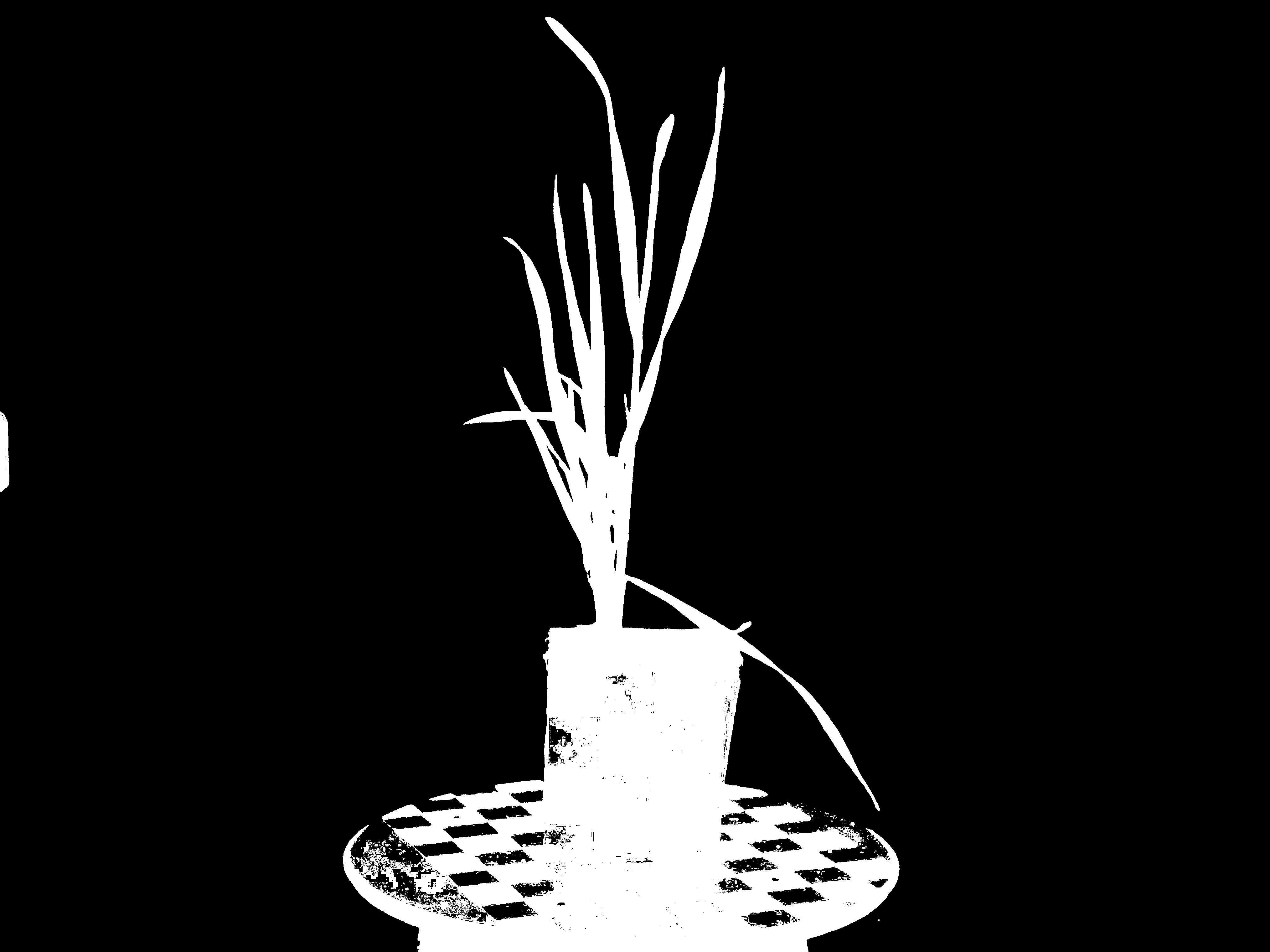}
		}
		\subfloat[\centering Camera D (Binary)]{
			\includegraphics[width=0.23\textwidth]{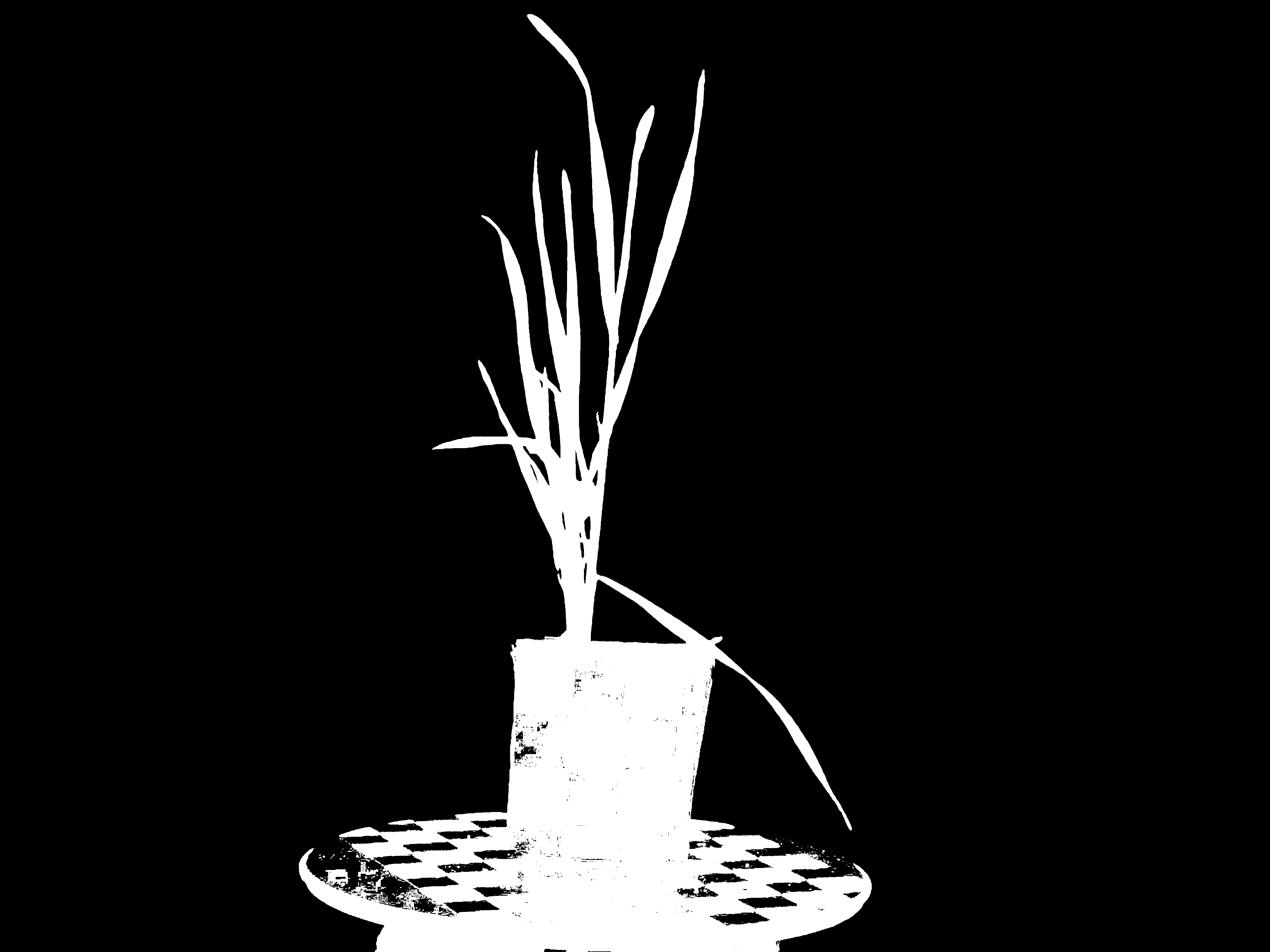}
		}
		\caption[Images from different viewpoints.]{Images captured from four different perspectives by cameras A through D (top row) and the corresponding binary images (bottom row). Camera A represents the highest vantage point, with each subsequent camera (B to D) capturing from progressively lower perspectives. The binary images (bottom row) show regions where no features will be extracted in black.}
		\label{fig:images-example}
	\end{figure*}

	Figures \ref{fig:images-example} (a)-(d) show how different perspectives on the vertical axis were captured by the four cameras on the photogrammetry rig. Figures \ref{fig:images-example} (e)-(h) show a series of binary images corresponding to the same set of images. To generate these binary masks, the original RGB images are first converted to the LAB color space, which separates lightness from color information. The masking process involves thresholding according to retaining pixels only if their \(b\)-channel has a value of 80 or higher and if their \(a\)-channel  value is below 140. This binary enhances the efficiency and accuracy of the feature extraction step in the 3D reconstruction. The threshold values have been optimized for stable indoor lighting conditions but may need adjustment under varying lighting environments.
	
	\subsection{Point cloud construction}
	
	A point cloud is defined as a set \(\{ P_i\}_{i = 1,\ldots,n}\), where each point \(P_i\) represents a vector detailing its spatial location in Cartesian  coordinates \((x,y,z)\) and additional information such as color, reflected light, or surface orientation. Generally, each point is a vector \(P_i = (x_i,y_i,z_i, f_{i,1},\ldots,f_{i,j})\), where \(j\) is the number of features attached to the point.
	
	For the point cloud construction we use the COLMAP software \cite{schoenberger2016sfm}. Its GUI allows users to visually inspect and interact with reconstructed models, whereas the command-line interface offers scriptability for automated workflows and parameter variations. For data storage and processing, COLMAP uses an SQLite database to store image paths, features, feature matches, and camera information/parameters. The reconstruction process then utilizes this database to generate a 3D model, saving it in a specified output folder as three binary files: one each for cameras, images, and 3D points. These files can be subsequently accessed through the COLMAP GUI for viewing and analyzing the reconstruction. The point clouds generated are in PLY format (Polygon File Format).
	
	We created a script that interfaces with COLMAP to automate the process of 3D reconstruction by sequentially executing COLMAP commands. Using this method the following steps are performed to construct the point cloud:
	\begin{enumerate}
		\item Calling the COLMAP feature extractor
		\item Find correspondences between the features across various images
		\item Execute the COLMAP mapper for a sparse 3D model
		\item Perform image undistortion
		\item Dense reconstruction running the patch match stereo algorithm
		\item Perform stereo fusion, integrating depth maps into the 3D model
		\item Export the model into PLY format
	\end{enumerate} 
    Details about the parameters used for each of these commands can be found on the Github \cite{githubrepo}.
	
	Table \ref{tab:reconstruction-times} shows the hardware specifications of two distinct systems used for 3D reconstruction tasks in this work, as well as the reconstruction time per point cloud on these systems. System 2 is hosted by the Digital Research Alliance of Canada (DRAC) and served as the primary platform for the 3D reconstruction computation in this work.
	
	\begin{table}[h]
		\centering
		\begin{tabular}{lll} 
			& \textbf{System} 1 & \textbf{System 2} \\
			\hline CPU & AMD 1920X 12-Core Processor @ 3.50 GHz & Intel Xeon(R) CPU @ 2.50GHz \\
			\hline RAM & $32 \mathrm{~GB}$ & $40 \mathrm{~GB}$ \\
			\hline GPU & NVIDIA Geforce RTX 3090 & NVIDIA Tesla V100 GPU \\
			  & Compute capability 5.0 & Compute capability 6.0 \\
			& 10496 CUDA cores & 5120 CUDA cores \\
			& Peak memory bandwidth $80 \mathrm{~GB} / \mathrm{s}$ & Peak memory bandwidth $549 \mathrm{~GB} / \mathrm{s}$ \\
			\hline GPU device memory & $24 \mathrm{~GB}$ & $12 \mathrm{~GB}$ \\
			\hline PCIe bus & v3.0 x16 $(8.0 \mathrm{GT} / \mathrm{s})$ & v3.0 x16 $(8.0 \mathrm{GT} / \mathrm{s})$ \\
			\hline C++ compiler & Microsoft Visual Studio &  GCC C++ compiler \\
			& C++ 2015 compiler &  \\
			\hline CUDA compiler & nvcc & nvcc \\
			\hline Operating System & Windows 10 & Ubuntu-22.04.3 \\
			\hline
			\hline Reconstruction Time & 120 minutes & 60-120 minutes \\
		\end{tabular}
		\caption[Hardware specifications for 3D Reconstruction systems]{Comparison of hardware specifications between two systems used for 3D reconstruction tasks in this work.}
		\label{tab:reconstruction-times}
	\end{table}
	
	\subsection{Point cloud processing}
	
\begin{figure}[h]
		\centering
		\includegraphics[width=0.9\linewidth]{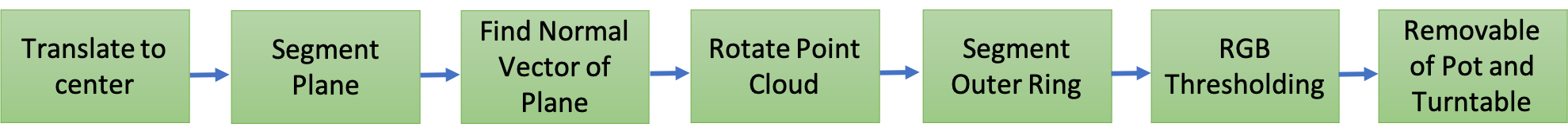}
		\caption{Flowchart of point cloud pre-processing steps.}
		\label{fig:processing-steps}
	\end{figure}

	To process the point cloud data, we utilize the Open3D library in Python, a toolset designed for the handling and analysis of 3D data \cite{Zhou2018}. Figure \ref{fig:processing-steps} shows a flowchart detailing the point cloud processing steps, each of which are described in more detail below.
	
\begin{figure}[h]
        \centering
		\begin{subfigure}{0.4\textwidth}
			\centering
			\includegraphics[width=0.9\linewidth]{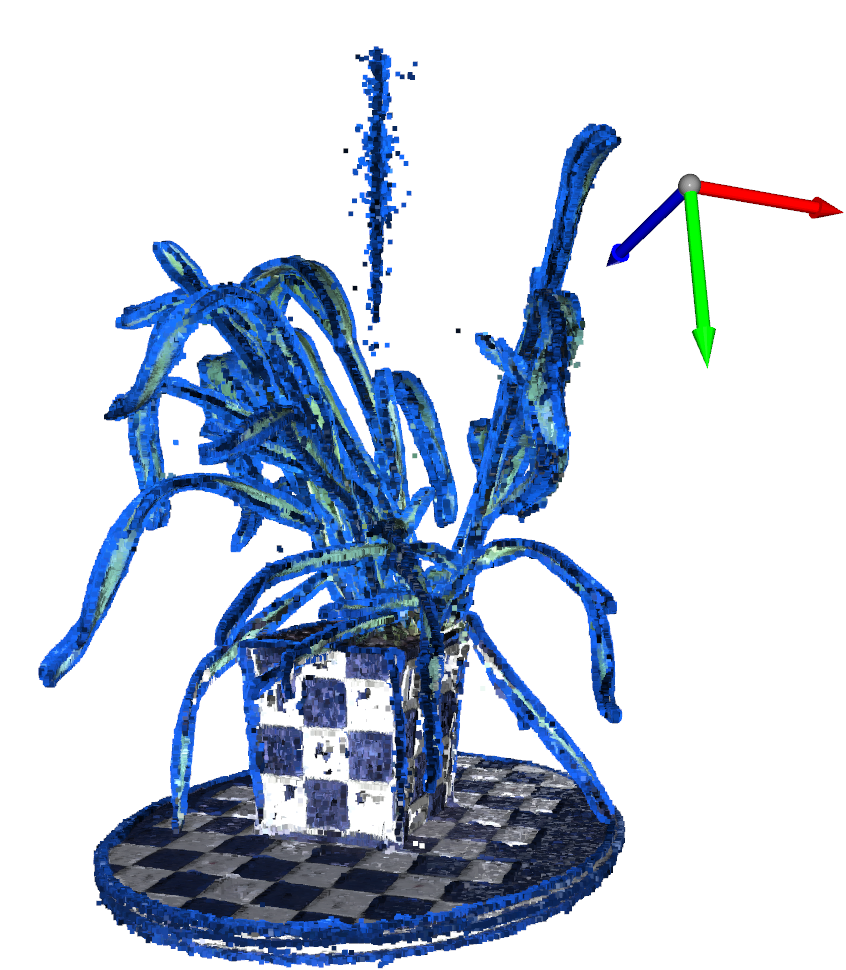}
		\end{subfigure}%
        \hspace{2cm}
		\begin{subfigure}{0.4\textwidth}
			\centering
			\includegraphics[width=0.8\linewidth]{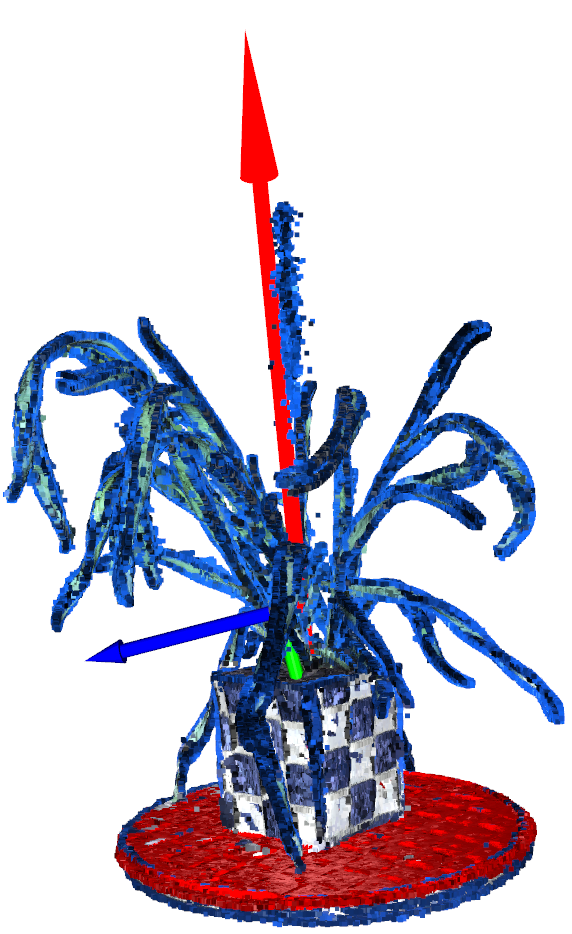}
		\end{subfigure}%
		
		\caption[Visualization of plane segmentation process]{(Left) Point cloud before alignment, the origin of the coordinate system lies clearly outside the object and is oriented in no particular relationship to the real-world object. (Right) The same point cloud after alignment. The point cloud origin lies in the center of the pots top surface and axes are aligned such that the pot surface coincides with the \(XY\)-plane and the \(Z\)-axis aligns with the normal of the turntable (red arrow, exaggerated for visibility). The part of the point cloud that represents the turntable used for this alignment is also highlighted in red.}
		\label{fig:aligning-point-cloud}
	\end{figure}

	For further processing, the point cloud must be centered and its orientation aligned with the positive \(XY\)-plane. This is done in the three steps: centering the point cloud, rotating the point cloud to align positive \(z\)-direction with the normal of the turntable plane, and finally translating the points such that the turntable plane aligns with \(z = 0\). For the first step, we subtract the average position of all points from each point. For the second and third step, we first identify the plane of the turntable inside the point cloud. We use an approach based on the RANSAC algorithm (see Algorithm \ref{alg:RANSAC}) to identify the best fit plane by maximizing the number of inliers within a specified distance threshold. Upon concluding the algorithm returns the parameters of the identified plane alongside its inliers, signifying a successful segmentation of a significant planar surface from the point cloud. Given the plane equation \(ax + by + cz + d = 0\), the normal vector of the turntable can be directly obtained as \(n = [a,b,c]\). We then align the normal vector with the \(Z\)-axis. Figure~\ref{fig:aligning-point-cloud} shows the plane segmentation process applied to a point cloud of a wheat plant taken 35 days post-planting. As can be seen in the image on the right, the point cloud has been translated to the center and aligned (as noted by the axis), with the red points indicating  the inliers that coincides with the turntable. The red arrow illustrates the direction of the normal vector relative to the segmented plane, showing the orientation of the turntable. Finally, the base of the point cloud is adjusted, so that $z=0$ is the plane of the turntable.
	
	\begin{algorithm}[h]
        \footnotesize
		\caption{Plane Segmentation in a Point Cloud}\label{alg:RANSAC}
		\begin{algorithmic}[1]
			\Function{SegmentPlane}{$pointCloud$, $distanceThreshold$, $maxIterations$}
			\State $bestPlane \gets$ null
			\State $bestInliers \gets$ empty set
			\For{$i \gets 1$ to $maxIterations$}
			\State $sampledPoints \gets$ Randomly sample 3 points from $pointCloud$
			\State $plane \gets$ Fit a plane to $sampledPoints$
			\State $inliers \gets$ empty set
			\ForAll{$point$ in $pointCloud$}
			\If{Distance from $point$ to $plane$ $<$ $distanceThreshold$}
			\State Add $point$ to $inliers$
			\EndIf
			\EndFor
			\If{Size of $inliers$ $>$ Size of $bestInliers$}
			\State $bestPlane \gets plane$
			\State $bestInliers \gets inliers$
			\EndIf
			\EndFor
			\State \Return $bestPlane$, $bestInliers$
			\EndFunction
		\end{algorithmic}
	\end{algorithm}
	
\begin{figure}[h]
		\begin{subfigure}[t]{0.46\textwidth}
			\centering
			\includegraphics[width=0.55\linewidth]{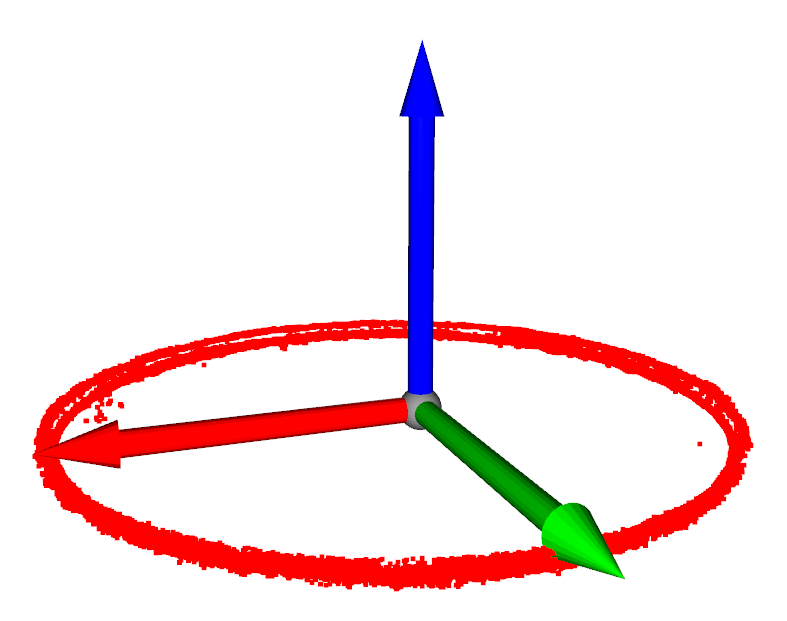}
		\end{subfigure}%
		\hfill
		\begin{subfigure}[t]{0.46\textwidth}
			\centering
			\includegraphics[width=0.6\linewidth]{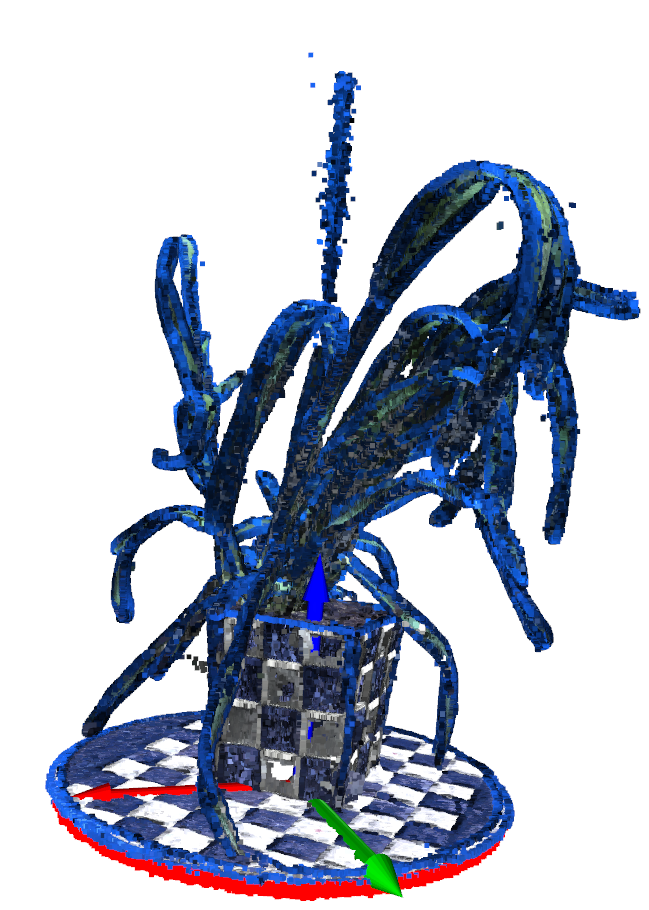}
		\end{subfigure}%
		
		\caption[Ring segmentation of the plant point clouds]{(Left) Segmented outer ring of the turntable used for scaling of the point cloud to real-world units. (Right) The complete point cloud view post-segmentation, showing the position of the point cloud and segmented turntable ring highlighted in red.}
		\label{fig:outer-ring-segmentation}
	\end{figure}

	The point cloud resulting from COLMAP is not scaled yet, i.e., the values of the \((x,y,z)\)-coordinates do not correspond to laboratory distances. To establish the scale of the point cloud we segment the outer points of the turntable plane, for which we know its actual diameter. To segment the ring of the turntable we calculate the radial distance of points with a small value for \(z\) from the center of the point cloud. By constructing a histogram of these radial distances, we can identify the most common radial distance, which coincides with the area where the turntable ring is most densely populated. This method (see Algorithm \ref{alg:outer-ring-segmentation} for details) allows for the accurate segmentation of the outer ring of the turntable. Figure \ref{fig:outer-ring-segmentation} shows the segmented outer ring of the turntable.

	\begin{algorithm}[h]
        \footnotesize
		\caption{Segment Ring Points}\label{alg:outer-ring-segmentation}
		\begin{algorithmic}[1]
			\Function{segment\_ring\_points}{pts, num\_bins, z\_min, z\_max, tolerance}
			
			\State filtered\_pts $\gets$ pts[(pts[:, 2] > z\_min) \, \& \, (pts[:, 2] < z\_max)]
			\State r $\gets$ sqrt(filtered\_pts[:, 0]$^2$ + filtered\_pts[:, 1]$^2$)
			\State hist, bin\_edges $\gets$ histogram(r, bins=num\_bins)
			\State max\_bin\_index $\gets$ argmax(hist)
			\State peak\_radius $\gets$ (bin\_edges[max\_bin\_index] + bin\_edges[max\_bin\_index + 1]) / 2
			\State ring\_indices $\gets$ where((r $\geq$ peak\_radius - tolerance) \& (r $\leq$ peak\_radius + tolerance))
			\State \Return ring\_indices
			\EndFunction
		\end{algorithmic}
	\end{algorithm}

\begin{figure}[h]
		\begin{subfigure}[t]{0.46\textwidth}
			\centering
			\includegraphics[width=0.75\linewidth]{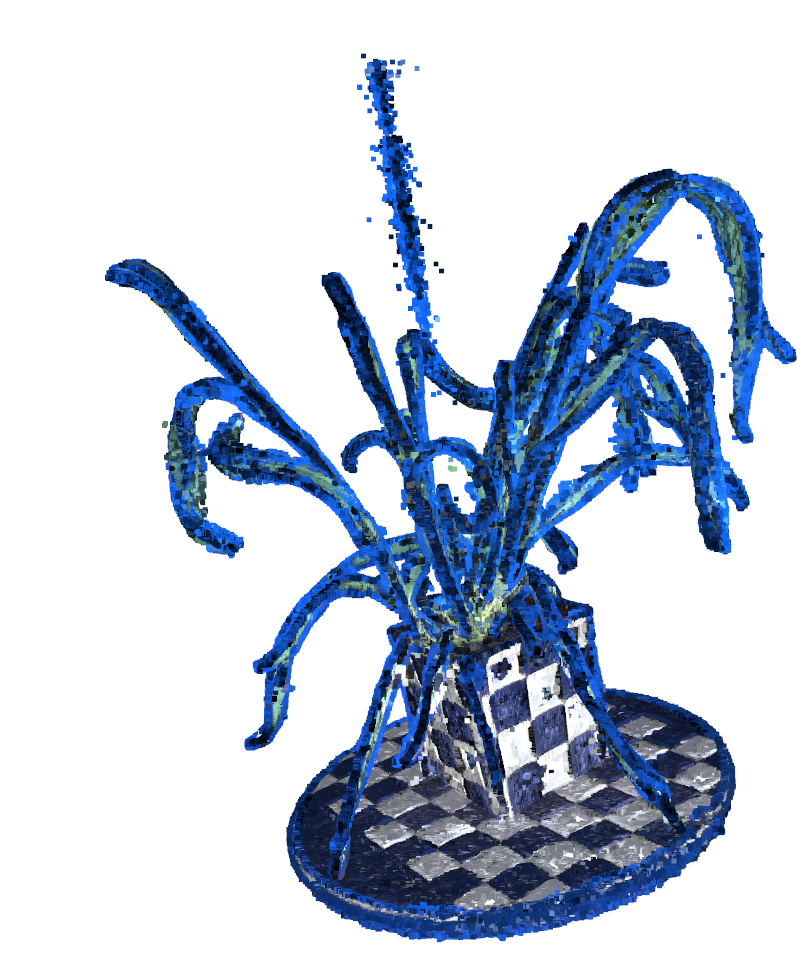}
		\end{subfigure}%
		\hfill
		\begin{subfigure}[t]{0.46\textwidth}
			\centering
			\includegraphics[width=0.7\linewidth]{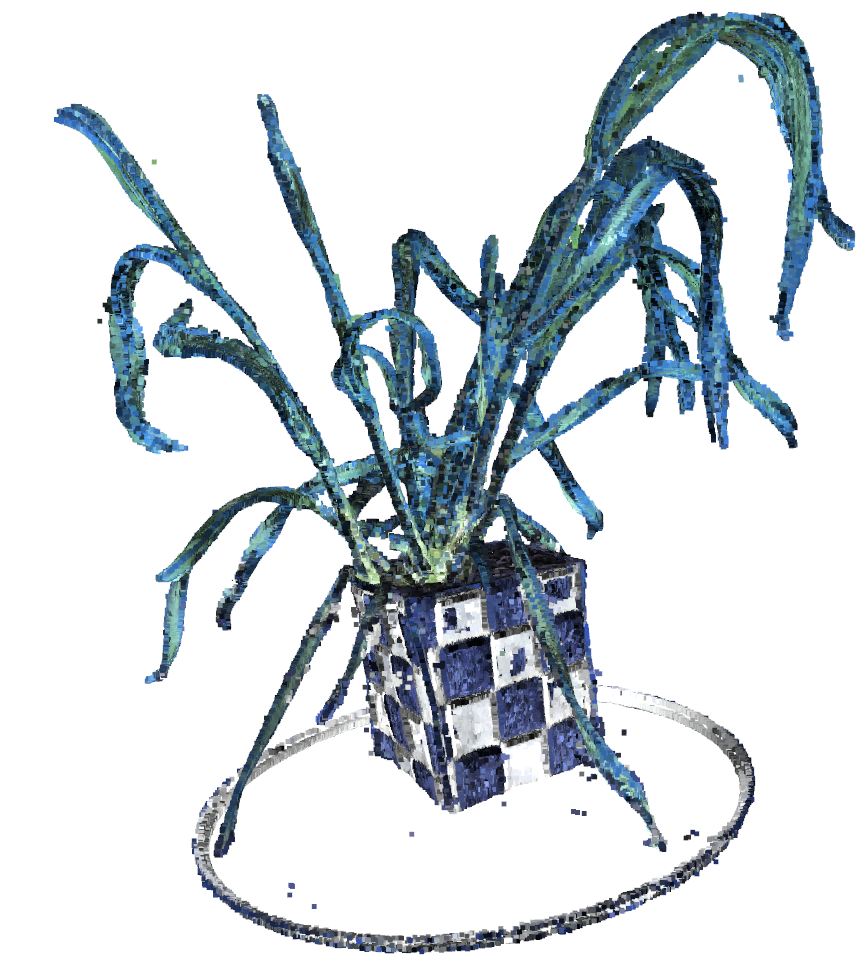}
		\end{subfigure}%
		
		\caption[RGB thresholding of the point cloud data]{(Left) Unprocessed point cloud showing blue artifacts around the 3D model of the object. (Right) Processed point cloud after applying RGB thresholding.}
		\label{fig:blue-removal}
	\end{figure}

\begin{figure}[h]
		\centering
		\subfloat{
			\includegraphics[width=0.3\textwidth]{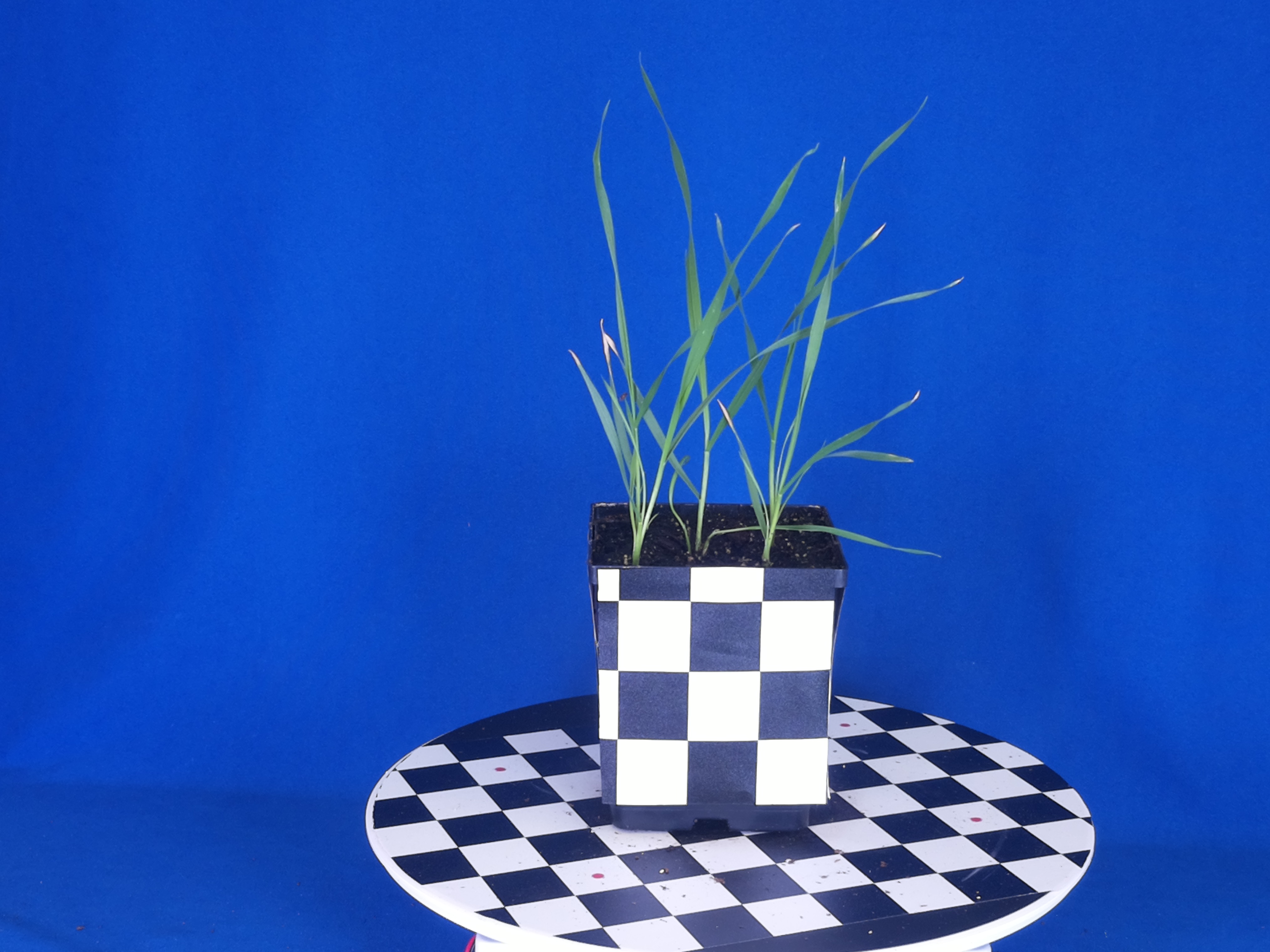}
		}
		\subfloat{
			\includegraphics[width=0.3\textwidth]{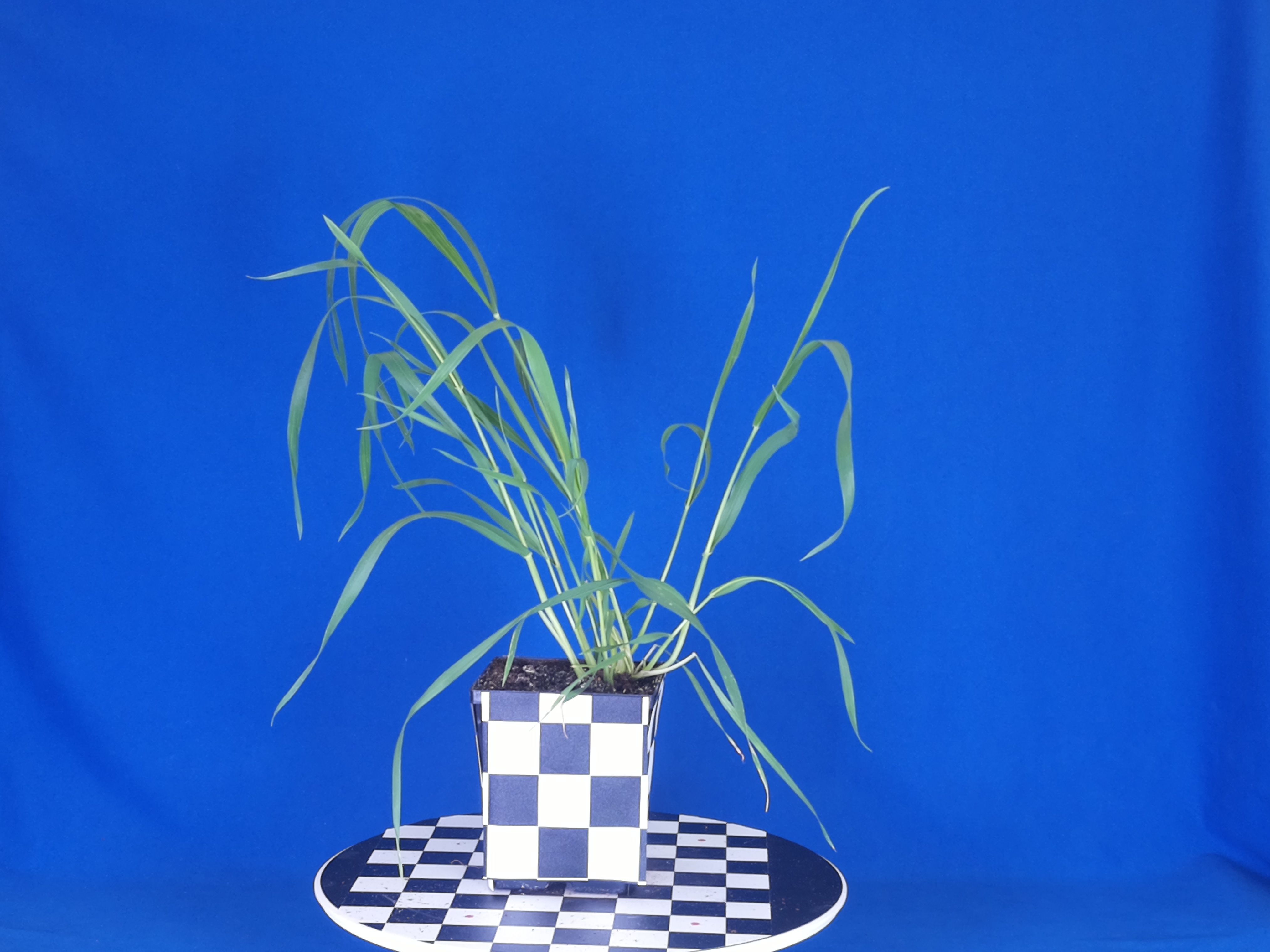}
		}
		\\ 
		\subfloat{
			\includegraphics[width=0.3\textwidth]{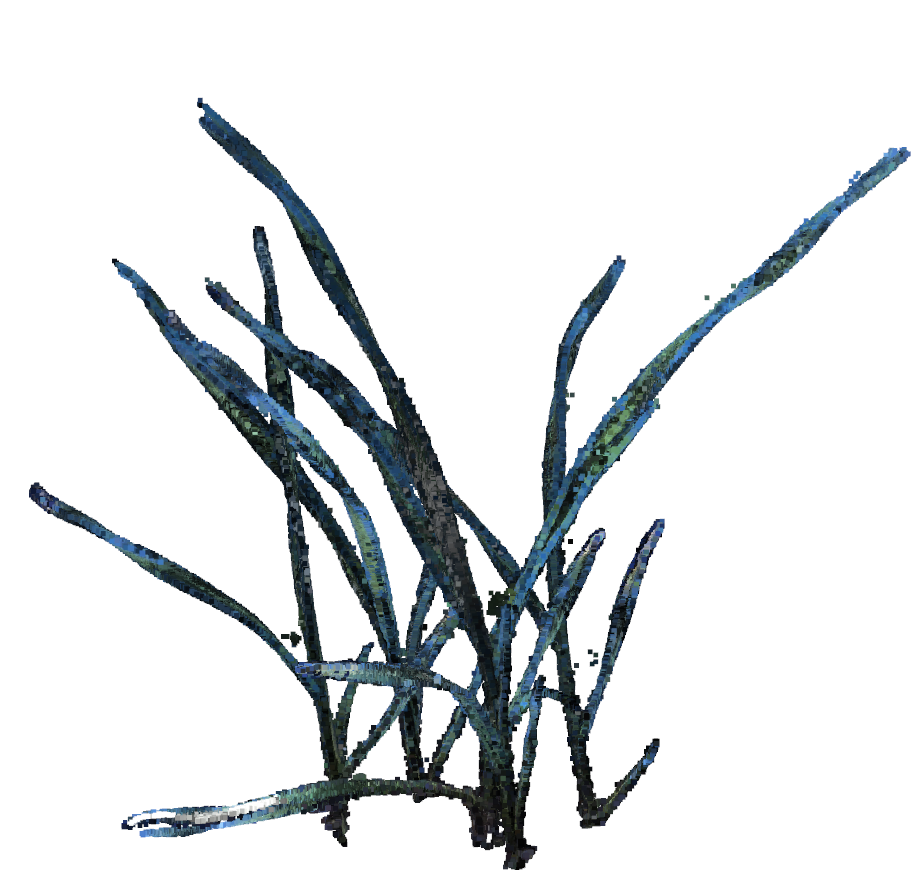}
		}
		\subfloat{
			\includegraphics[width=0.3\textwidth]{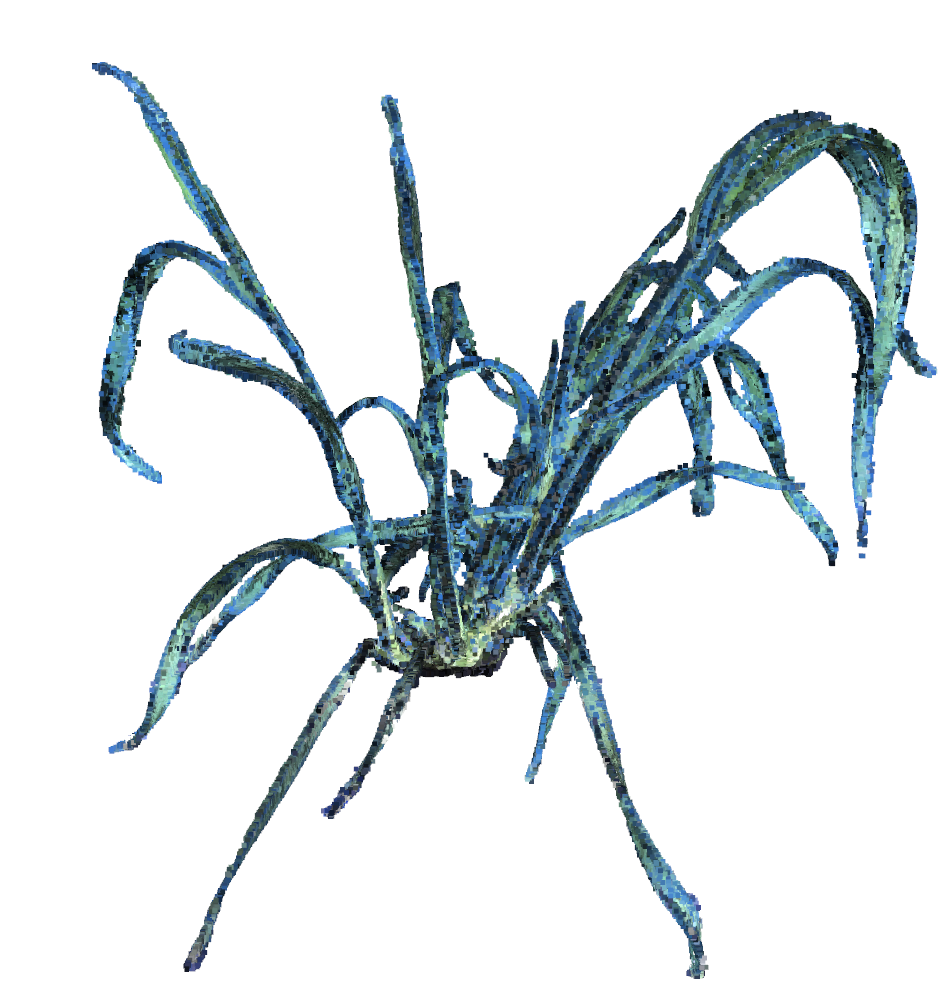}
		}
		\caption[Visual of the wheat growth stages]{Examples of fully processed point clouds (bottom) and sample photographs (top) used in their construction. These are shown for the planophile Alsen wheat genotype at 14 days (left) and 35 days (right) after planting.}
		\label{fig:Alsen-reconstruction}
	\end{figure}

	While the blue uniform backdrop helps with the 3D reconstruction, it also introduces additional blue points at the edges of the reconstructed object. We use color thresholding to eliminate those superfluous points, details are given in Algorithm \ref{alg:blue-removal}. The effect of the algorithm is illustrated in Figure \ref{fig:blue-removal}. In the final step we use Meshlab to remove the pot and turntable sections from the point cloud thus isolating the plant. This can be done easily as the dimensions of the pot and the turntable are known after aligning and scaling the point cloud. Figure \ref{fig:Alsen-reconstruction} shows a fully processed  point cloud.
	
	\begin{algorithm}[h]
        \footnotesize
		\caption{Filter RGB Points}\label{alg:blue-removal}
		\begin{algorithmic}[1]
			\Function{FilterNonPoints}{$points$, $colors$, $lower\_threshold$, $upper\_threshold$}
			\State $outside\_threshold\_indices \gets \Call{All}{(colors < lower\_threshold) \text{ OR } (colors > upper\_threshold), axis=1}$
			\State $filtered_points \gets points[outside_threshold_indices]$
			\State $filtered_colors \gets colors[outside_threshold_indices]$
			\State \Return $(filtered_points, filtered_colors)$
			\EndFunction
		\end{algorithmic}
	\end{algorithm}
	
\clearpage
    
	\section{Measurements and Quantitative Analysis of Wheat Plants}\label{sec:measurements}
	\subsection{Data collection}\label{sub:data-collection}
	To explore the capabilities of our photogrammetry system, we grew wheat plants (\emph{Triticum aestivum}) belonging to six genotypes (see Table \ref{tab:genotypes}). For each genotype ten pots were seeded with five seeds each. The  60 pots were kept in a growth chamber at 25 \degree C/day and 18 \degree C/night, with a 16-hour photoperiod, and relative humidity at 60\%. Planting was staggered over the duration of 4 days from June 13 to June 16, 2023. The seedlings were imaged with the photogrammetry system 14 days and 35 days after their respective planting days.  At each timepoint, the plants were taken out of the growth chamber, imaged for an estimated 25 minutes per plant, and then returned to the chamber. The setup facilitated automatic image acquisition from each camera and movement of the rotary table after every four sets of images were captured. Due to the  autofocus capability of the cameras, there was no need for further adjustments. 
	
	\begin{table}[h]
		\centering
        \normalsize
		\begin{tabular}{ll}
			\hline
			\textbf{Genotype} & \textbf{Canopy Architecture Rating} \\
			\hline
			\text{Alsen} & $ \text{10 (extreme planophile)} $ \\
			\hline
			\text{AAC Brandon} & \text{10 (extreme planophile)} \\
			\hline
			\text{CDC Teal} &  \text{7 (planophile)}\\
			\hline
			\text{Chara} & \text{3 (erectophile)} \\
			\hline
			\text{Gladius} & \text{1 (extreme erectophile)} \\
			\hline
			\text{Kukri} & \text{5 (neutral)} \\
			\hline
		\end{tabular}
		\caption[Genotypes imaged with the system]{Genotypes imaged and their manullay assigned canopy architecture rating that ranges on a scale from 1 to 10.}
		\label{tab:genotypes}
	\end{table}

After the imaging session of Day 14, the plants were selectively thinned, leaving only the most robust specimen in each pot for measurements on Day 35. Table \ref{tab:imaging-settings} outlines the configuration of the photogrammetry rig used in both these sets of experiments. The turntable was set to rotate at 10\degree\, to guarantee substantial overlap among the images captured. To capture the entirety of each plant while maintaining a consistent background, the distance between the plant and the rig was kept within 1-2 meters. 
    
	\begin{table}[h]
		\centering
        \normalsize
		\begin{tabular}{ll}
			\hline
			\textbf{Setting} & \textbf{Value} \\
			\hline
			\text{Turntable Angle} & $10^\circ$ \\
			\hline
			\text{Total Images} & 148 \\
			\hline
			\text{Camera Resolution} & \text{4084 x 3051} \\
			\hline
			\text{Focal Length} & \text{16 mm} \\
			\hline
			\text{Aperture Setting} & \text{f/5.6} \\
			\hline
			\text{ISO Setting} & \text{ISO-143} \\
			\hline
			\text{Distance to Subject} & \text{1-2 m} \\
			\hline
			\text{Total Imaging Time} & \text{25 Minutes} \\
			\hline
			\text{Total Size on Disk} & \text{297 MB} \\
			\hline
			\text{Pause Before Image Capture} & \text{6 seconds} \\
			\hline
		\end{tabular}
		\caption[Settings of the photogrammetry rig for wheat image capture]{Imaging parameters used for capturing wheat images.}
		\label{tab:imaging-settings}
	\end{table}
	
	From these images, a total of 120 point clouds were generated: 60 multiple-plant point clouds from Day 14 and 60 single-plant point clouds from Day 35. Reconstructions were performed on a remote server at the Digital Research Alliance Canada (DRAC) to speed up processing. Each reconstruction took between 60-120 minutes to complete. Point clouds were generated using the COLMAP interface script as outlined in Section \ref{sec:software}. Each point cloud then underwent processing as detailed in the previous section. Figure \ref{fig:Alsen-reconstruction} shows the growth stages of the Alsen wheat genotype (planophile) at days 14 and 35 after planting in ordinary images and as the reconstructed point cloud data.
	
	\subsection{Semi-Automated Segmentation Pipeline}
	Accurately isolating plant parts within a 3D point cloud is essential for morphological analyses (e.g., leaf angles, biomass estimation). The day-14 wheat datasets contain multiple seedlings within a single pot, making it challenging to distinguish individual plants from each other. Our semi-automated graph segmentation pipeline efficiently partitions the dense point clouds into separate seedlings.
	
	The pipeline consists of the following steps:
	\begin{enumerate}
		\item A voxel-based downsampling reduces point density while preserving overall geometry. We estimate normals to capture surface orientation characteristics and remove statistical outliers, yielding a sparser, cleaner dataset optimized for subsequent processing.
		
		\item To enhance the impact of vertical features, we apply anisotropic scaling to the \(z\)-axis before constructing a k-nearest neighbors graph. Pruning edges above a distance threshold yields connected components and a unique cluster label is assigned to each component.
		
		\item  Large or merged clusters are further subdivided by computing a distance matrix and forming a minimum spanning tree (MST) and cutting its largest edges (typically one or two cuts). This step isolates closely adjacent leaves and stems that would otherwise appear fused in the initial segmentation.
		
		\item Any cluster smaller than a specified size threshold is labeled as noise (-1). The pipeline also identifies particularly large clusters using statistical methods (percentile-based, standard deviation, or k-means).
		
		\item Users can export specific clusters for manual refinement in external tools such as Meshlab. Once manual edits are completed, the modified labels are integrated back into the point cloud data.
		
		\item A nearest neighbour search based on a \(k\)-d tree is used to transfer the refined labels from the downsampled point cloud back to the original, full-resolution cloud.
			
	\end{enumerate}
	
\begin{figure}[h]

	\centering
	\hspace{1cm}
	\includegraphics[clip, trim=900 210 800 120, height=6.0cm]{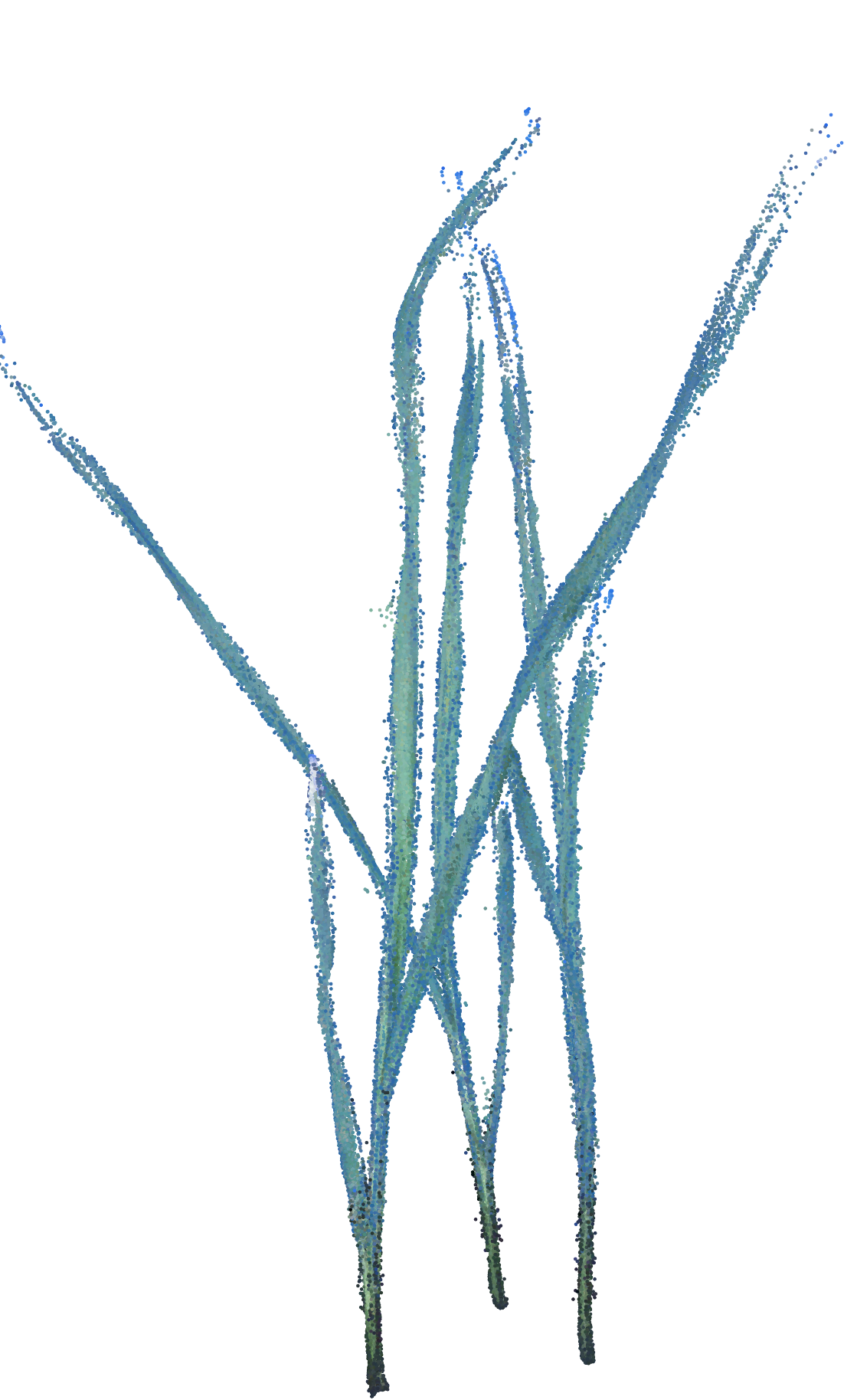}\hspace{0.1cm}
	\includegraphics[clip, trim=800 165 700 100, height=6.0cm]{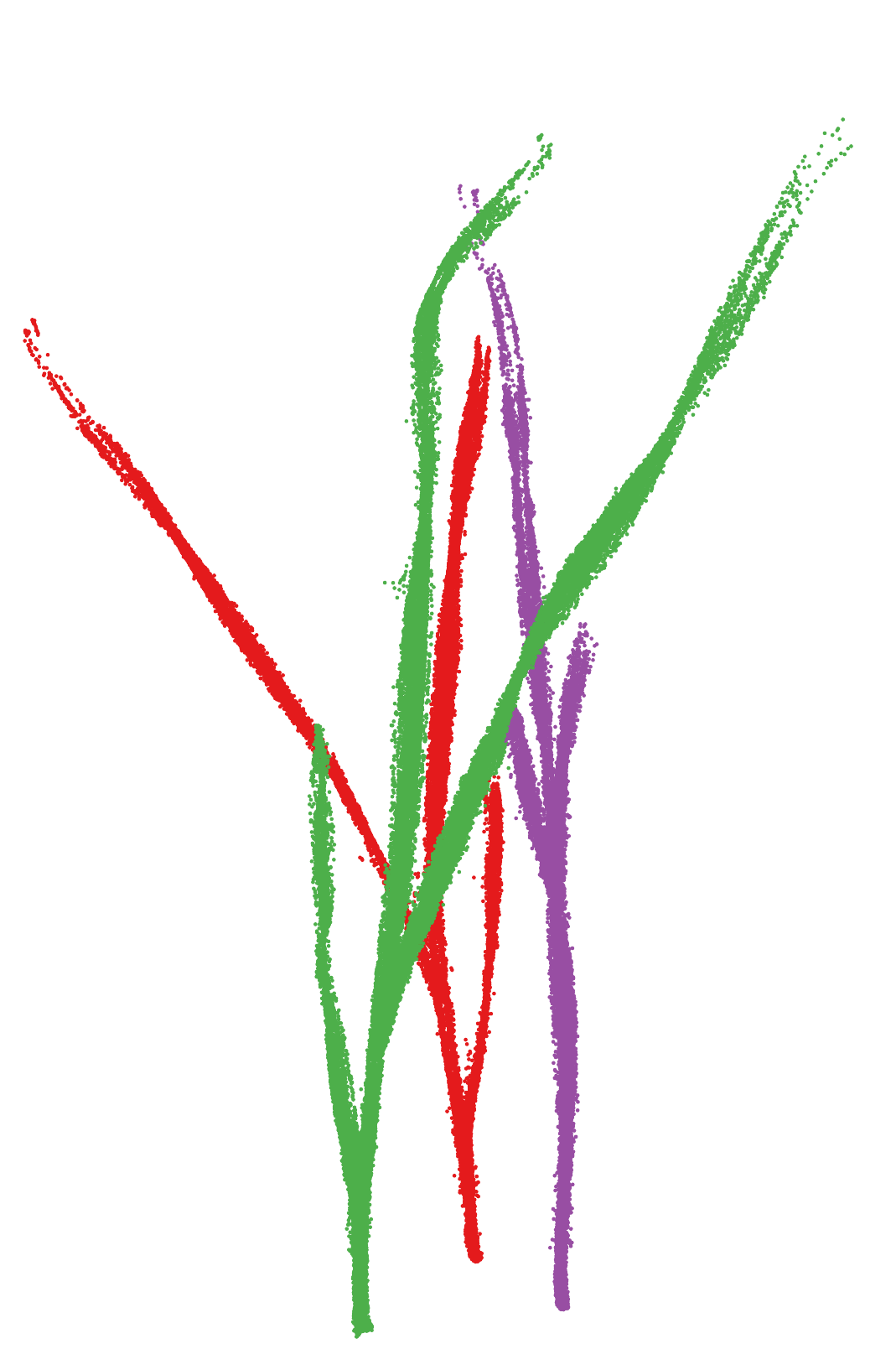}

	\caption[Pipeline segmentation]{Example of the segmentation pipeline on a point cloud of Day 14 Chara. (Left) It can be seen how the leaf of one of the plants visually crosses the stem of the other plant. (Right) Each plant has been distinctly labeled, indicated by the different colorings.}
	\label{fig:segmentation-pipeline-example}%
	\end{figure}

	Figure~\ref{fig:segmentation-pipeline-example} illustrates the pipeline's results on a day-14 wheat pot, demonstrating color-coded segmentation of individual seedlings and highlighting the overlapping region where two seedlings' branches overlap. By integrating graph-based clustering, MST refinement, and manual editing, this pipeline provides a robust yet flexible framework for segmenting dense plant point clouds containing multiple seedlings. With this pipeline we managed to create a dataset of 285 individually labeled wheat seedlings that can now be analyzed for trait extraction and phenotypic characterization.

	\subsection{Phenotypical Traits Extracted}
	From the point clouds of wheat obtained with our system, we subsequently extracted a variety  phenotypical features.  We focused on the following here: height, radius (i.e., the maximal extension of the plant from its central axis), convex hull and volume, ground cover projection, and angles between leaf and stem.
	All these metrics can be futher filtered, such that they apply to the whole plant or to parts of the plant. For example, it can be useful to calculate the convex hull for only the top 40\% of the plant (with respect to total height) or to horizontally slice the plant and apply these metrics to each slice individually. By developing further algorithms, more features can be extracted from the point cloud. The following is a short description and visualization of each of the aforementioned features.
	
	\subsubsection{Plant Height}~\\
	The plant height is measured from the z-coordinates of the lowest to the highest plant voxel after the processing described in the previous section was performed (see Fig.~\ref{fig:height-and-radial}).
	
	\subsubsection{Plant Radius}
\begin{figure}[h]
		\begin{subfigure}[t]{0.4\textwidth}
			\centering
			\includegraphics[width=0.9\linewidth]{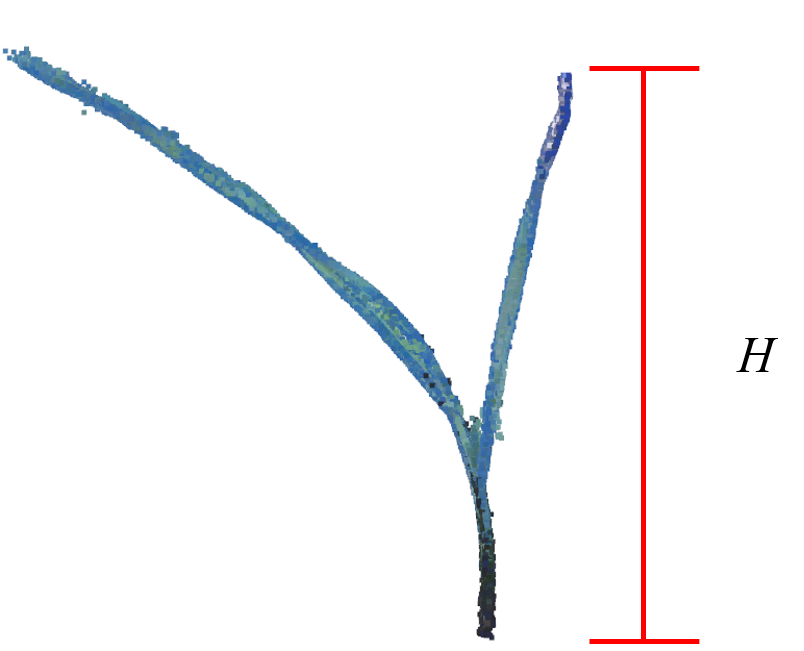}
		\end{subfigure}%
		\hfill
		\begin{subfigure}[t]{0.6\textwidth}
			\centering
			\includegraphics[width=1\linewidth]{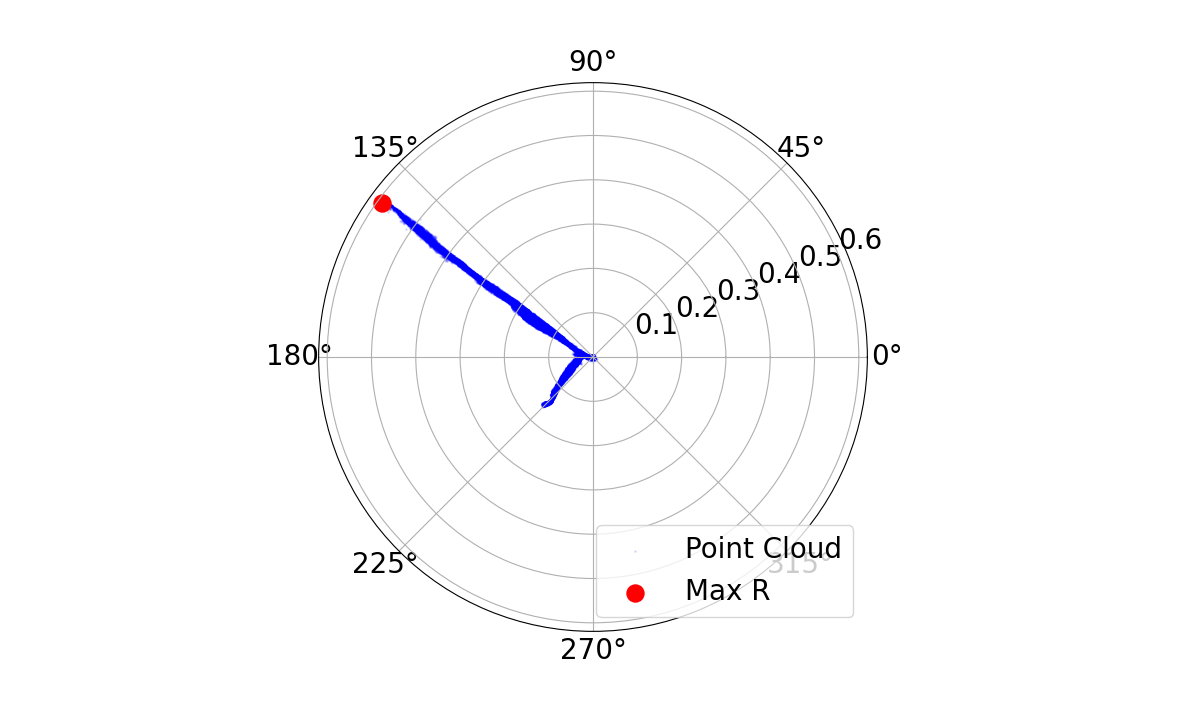}
		\end{subfigure}%
		\caption[Day 14 Brandon pot C4 cluster point cloud and polar plots]{Plots of one of Brandon wheat genotype seedling on day 14. (Left) The vertical height is defined by the largest \(z\)-coordinate of all voxels. (Right) Shows a polar plot of the same point cloud (projected onto the \(XY\)-plane), the plant radius is defined by the voxel with the largest radial distance to the plant center, marked in the plot by a red dot.}
		\label{fig:height-and-radial}%
	\end{figure}

~\\
	The plant radius is measured as the voxel that has the largest radial distance from the plant's central axis (as described in \ref{sec:software}). The height of the voxels has no bearing on this measurement. We illustrate the radius measurement in Fig.~\ref{fig:height-and-radial}, where the most distant voxel is highlighted by a red dot.
	
	\subsubsection{Convex Hull}
\begin{figure}[b]
        \centering
		\includegraphics[width=0.275\linewidth]{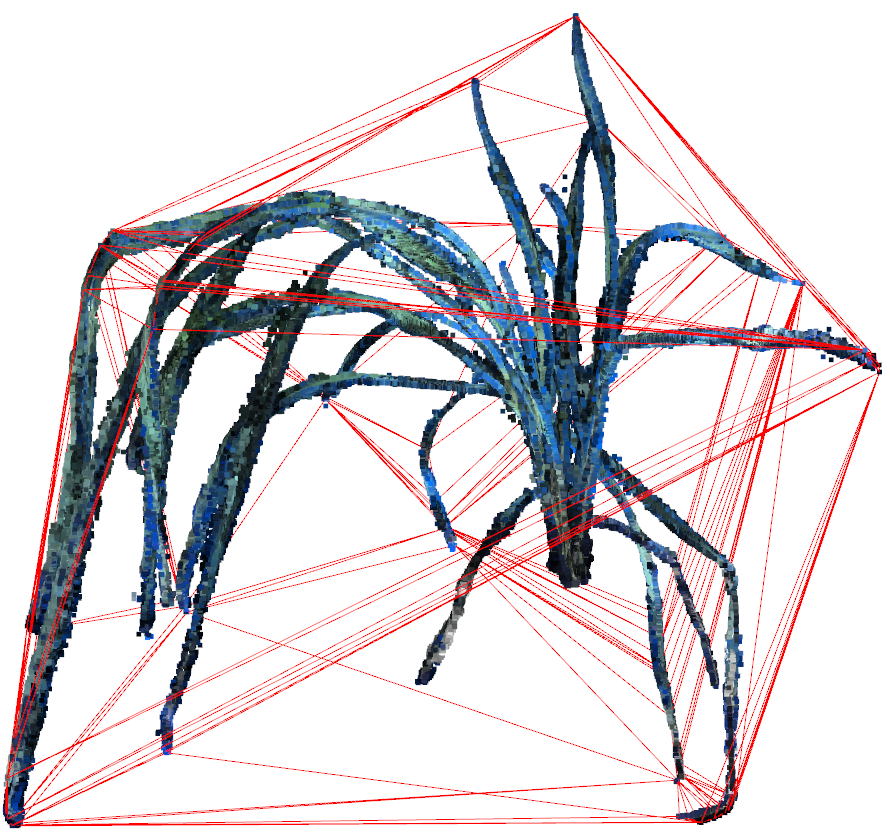}

		\caption[Brandon planophile visualization]{Visualization of the 3D point cloud for Brandon day 35 as the smallest convex volume encompassing the entire plant.}
		\label{fig:convex-hull}%
	\end{figure}

\begin{figure}[h]
		\centering
		\subfloat{
			\includegraphics[width=0.3\textwidth]{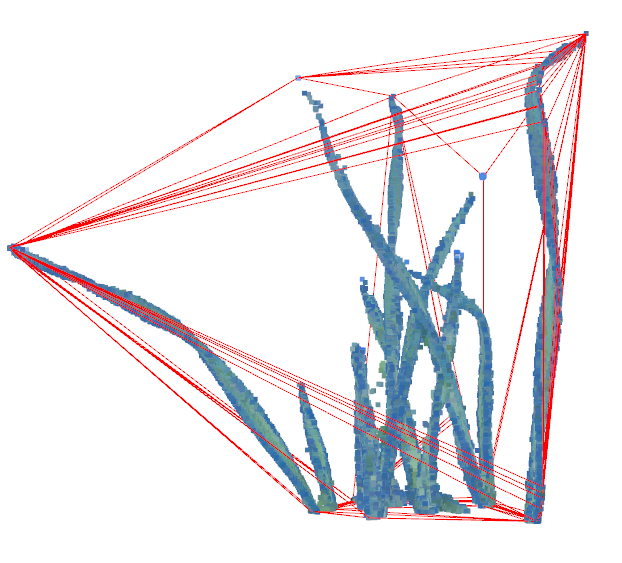}
		}
        \hspace{2cm}
		\subfloat{
			\includegraphics[width=0.3\textwidth]{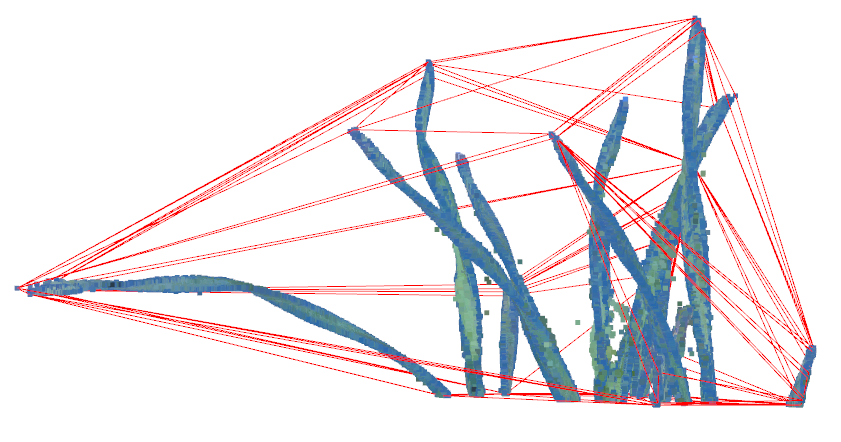}
		}
		\caption[Visualization of canopy volume at 60\%]{The canopy volume of the top 60\% of voxels for an example of the Kukri (Left, neutral canopy architecture) and Alsen (Right, extreme planophile canopy architecture) genotypes.}
		\label{fig:wheat_60_percent_canopy_volume}
	\end{figure}

~\\
	The convex hull of a point cloud is defined as the smallest possible convex volume that contains the entire point cloud. The convex hull has the advantages of being tighter around the plant than, say, the smallest possible cuboid, while still being clearly defined, easy to measure, and its volume easy to calculate. Figure \ref{fig:convex-hull} shows an example of a plant and its convex hull from different perspectives. To better distinguish canopy structures from each other we have also measured the convex hull of the top 60\% and 40\%, respectively, where voxels are discarded according to the value of their \(z\)-coordinate. This reduces the impact of the stem-length on the measurement, i.e., the volume of the canopy becomes decoupled from its height above the ground due to longer or shorter plant stems. (See Fig.~\ref{fig:wheat_60_percent_canopy_volume} for a visualization of the top 60\% convex hull). 
	
	\subsubsection{Area of Ground Cover Projection}
\begin{figure}[h]
		\centering
        \includegraphics[width=0.75\linewidth]{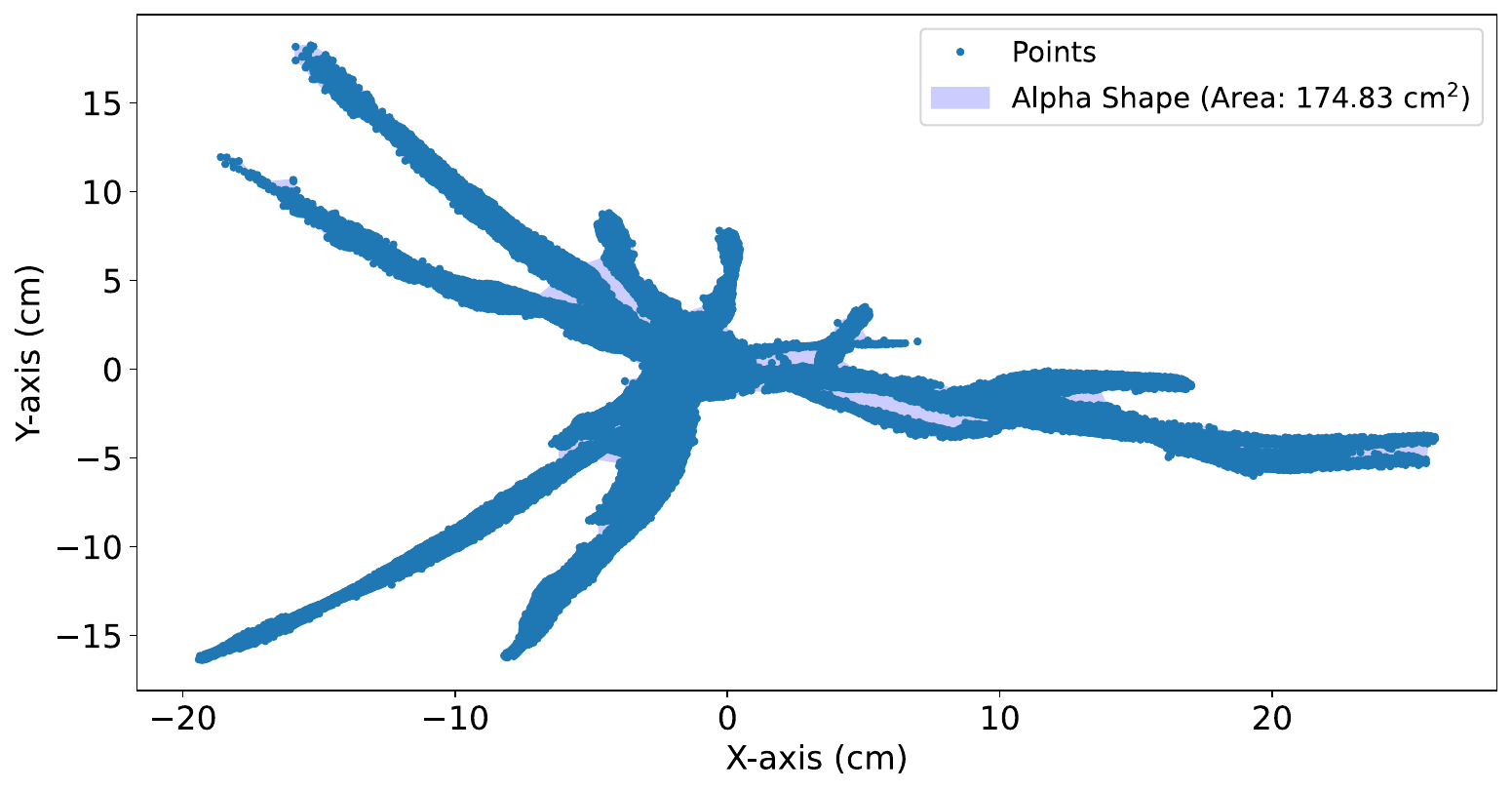}
		\caption[Brandon alpha-shape visualization]{Visualization of the plant voxels projected onto the \(XY\)-plane with corresponding alpha-shape to calculate the plant's canopy coverage.}
		\label{fig:alphashapes}%
	\end{figure}

~\\
	The ground cover projection is determined by orthogonally projecting the point cloud on the XY-plane (i.e. $z=0$) and is achieved by a 2D plot of the $x,y$ coordinates only. The area of this projection defines the ground coverage of the plant, i.e., the amount of ground that would be shadowed by the plant due to a far-away light source right above it. 
    The points in a point cloud are of course discrete, which makes defining and calculating the area encompassed by the projection non-trivial. To achieve this, we employed alpha-shapes~\cite{alphashapes} on the point cloud after flattening its z-axis. This shape can be considered as a generalization of the convex hull (\(\alpha=0\)) that tracks the area of the (now 2D) point cloud more precisely. We observed that a value of \(\alpha=100\) results in a reasonable ground cover projection. An example of a ground cover projection and area determination is shown in Fig.~\ref{fig:alphashapes}.

	\subsubsection{Leaf Angles}~\\	
	\begin{figure}
		\centering
		\includegraphics[width=0.45\linewidth]{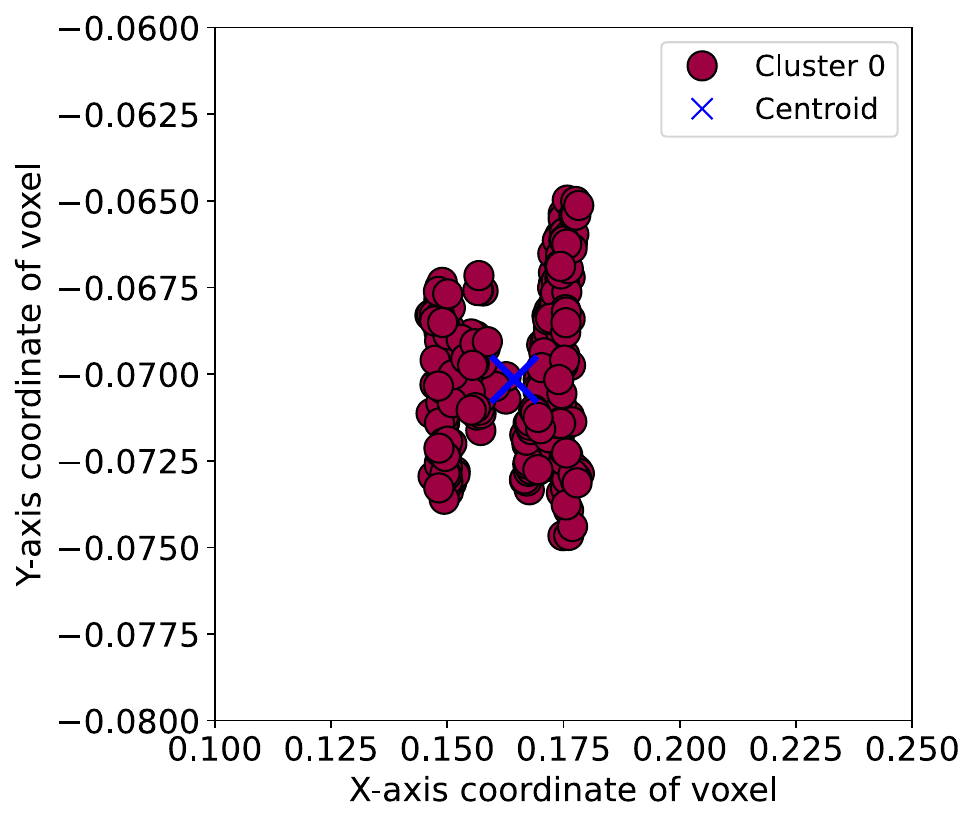}
		\includegraphics[width=0.45\linewidth]{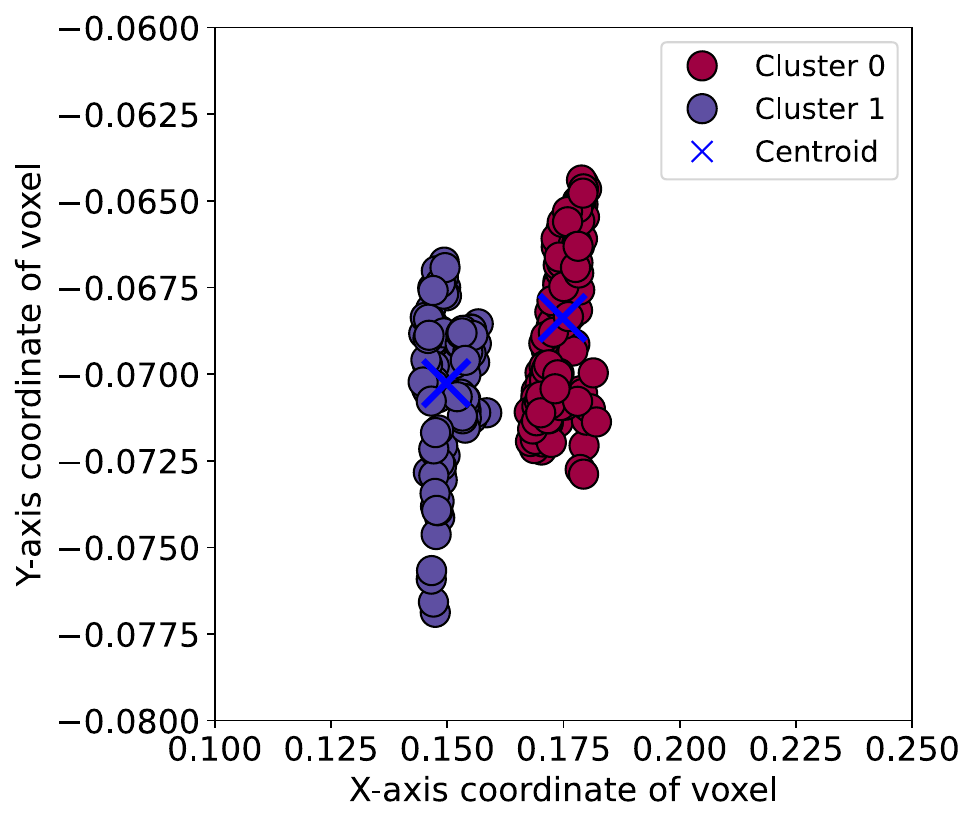}
		\caption{Two consecutive slices of a wheat plant showing how DBSCAN recognizes the splitting of a leaf off the main stem, by increasing the number of identified clusters by one. Each cluster's centroid is marked as a blue `x'.}
		\label{fig:consecutive-clusters}
	\end{figure}
	To determine the angle that a leaf forms with its stem, we perform the following individual steps on the point cloud:
	\begin{enumerate}
		\item The data is partitioned along the \(z\)-axis into \(n\) slices of equal thickness, where \(n\) is chosen such that the slices are sufficiently thin (in our experiments we chose \(n = 80\)).
		
		\item Within each slice, we apply the DBSCAN clustering algorithm~\cite{dbscan}. This algorithm discards outliers and groups the remaining points into clusters each representing a slice of a leaf or the stem of a plant. Note that when a plant "splits", i.e., when a leaf branches off the stem, the number of clusters in consecutive slices increases. See Fig.~\ref{fig:consecutive-clusters} for an example of this effect.
		
		\item For each slice and each cluster we determine its centroids.

		\item Having clusters in each horizontal slice, we establish adjacency links between slices \(i\) and \(i+1\) using two complementary matching methods:
		\begin{itemize}
			\item \emph{Distance-Based Bipartite Matching:} We compute an optimal one-to-one assignment between clusters in consecutive slices using the Hungarian algorithm to match clusters whose centroids lie within a specified distance threshold.
			\item \emph{Vertical/Trunk-Axis Matching:} A weighted cost function penalizes both horizontal offset $d_h$ and angular deviation $\theta_a$ from the estimated principal axis of the plant. The cost is formulated as \(\text{cost} = d_h + \beta \cdot \theta_a\), where \(\beta\) controls the relative importance of directional alignment.
		\end{itemize}
		
		\item To address disconnected subgraphs in the adjacency structure, we employ an incremental bridging technique, progressively relaxing the distance threshold until a single connected component emerges. 
		
		\item The algorithm selects the adjacency method that yields the lowest-cost path, making it robust to both vertical and slightly tilted plant. This creates a graph embedded in 3D space, as seen in Fig.~\ref{fig:leaf-angle-calculation}.
		
		\item We then construct a negative cost directed acyclic graph (DAG) resembling a "raindrop model" for stem extraction. This model works in several phases:

    \begin{figure}[t]
    \center

    \includegraphics[clip, trim=800 200 600 500, width=0.3\textwidth]{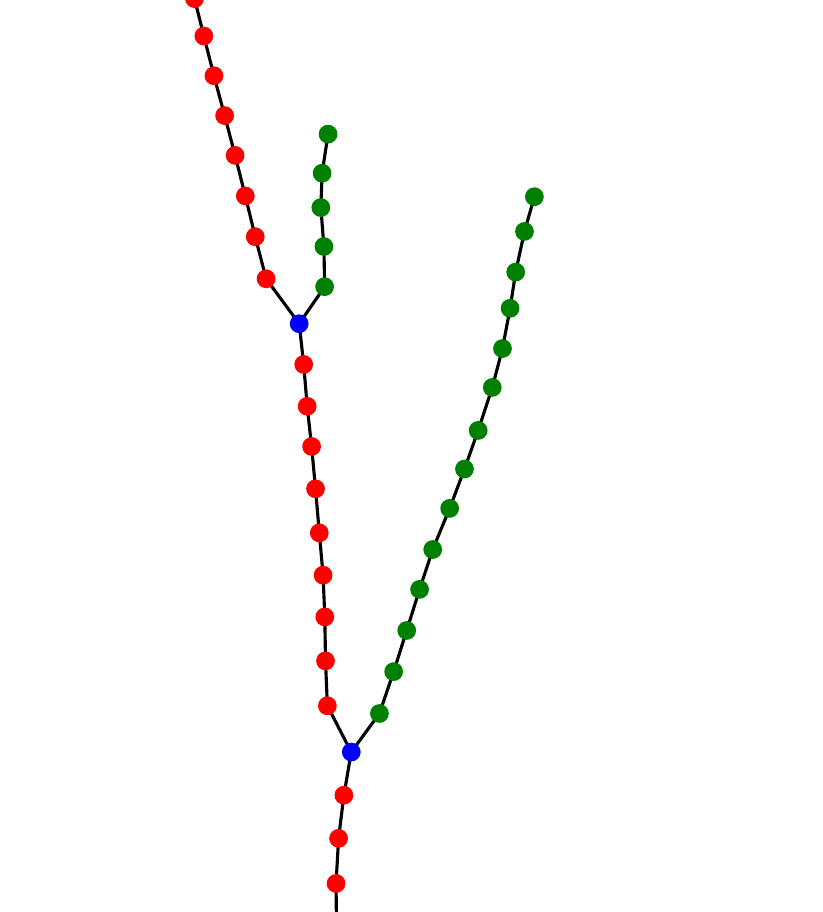}
    \hspace{0.0cm}
    \includegraphics[clip, trim=800 200 800 100, width=0.15\textwidth]{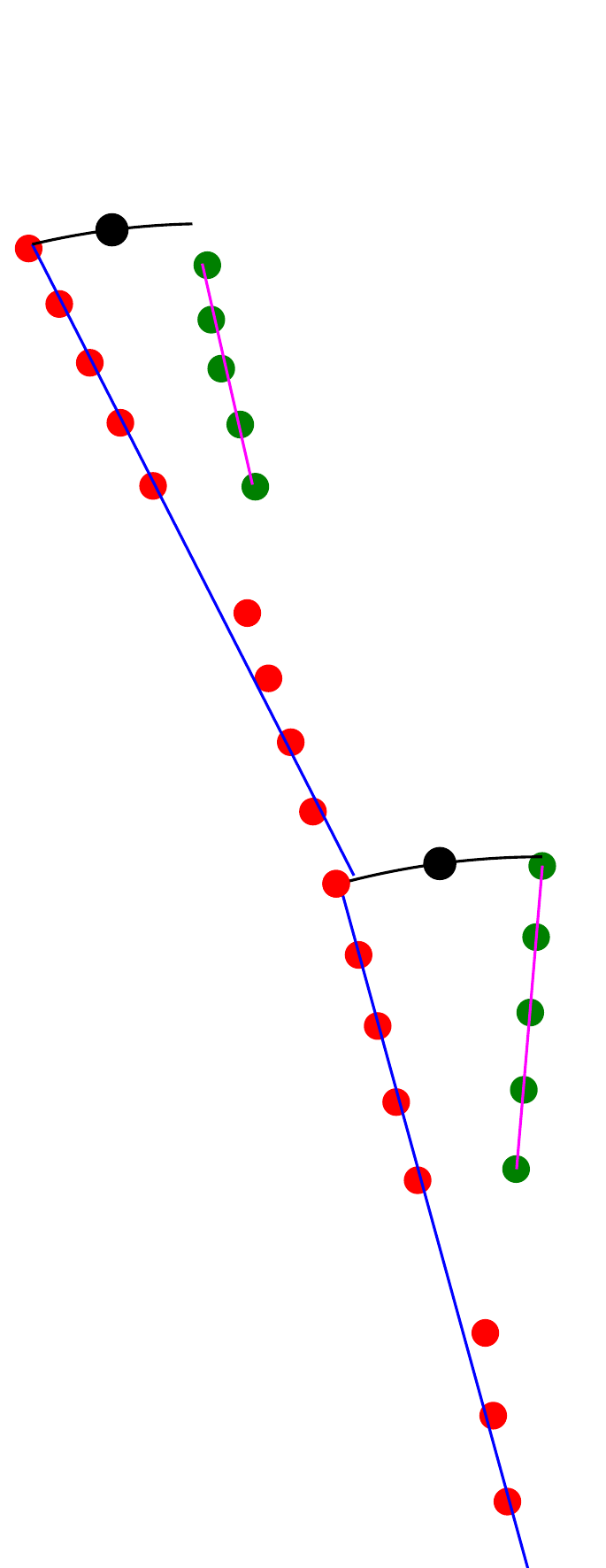}
    \caption[leaf angle calculations]{Visualization of the steps performed for leaf angle calculations. (Left) The 3D-embedded graph of connected centroids. Branching points of the plant can easily be identified as nodes of degree 3. (Right) The leaf angles are determined by calculating the angle between two regression lines through the centroids that belong to the stem and leaf, respectively.}
    \label{fig:leaf-angle-calculation}%
    \end{figure}
        
		\begin{itemize}
			\item \emph{Branch Point Identification:} First, we construct an undirected graph from our adjacency structure and identify potential branch points as nodes with degree $\geq 3$. Each branch point is assigned a branch count value indicating the number of branches (excluding the main trunk connections).
			
			\item \emph{DAG Construction:} Edges are directed from higher to lower z-coordinates to enforce top-to-bottom flow, with each edge $(u \rightarrow v)$ existing if and only if $z_v < z_u$. The edge cost combines multiple components \(\text{cost}_{edge} = -(\alpha \cdot d_h + \beta \cdot \theta_a + \gamma \cdot p_v + \delta \cdot b)\)
			where:
			\begin{itemize}
				\item \(d_h\) is the xy-plane displacement between centroids
				\item \(\theta_a = \arccos\left(\frac{\vec{v} \cdot \vec{axis}}{|\vec{v}|}\right)\) measures alignment with the principal axis
				\item \(p_v = (1 - \frac{\text{vertical\_diff}}{\text{vertical\_diff} + d_h})\) prioritizes more vertical connections
				\item \(b = (\text{branch\_count}_u + \text{branch\_count}_v)\) encourages the path to pass through detected branch off nodes
			\end{itemize}
			The parameters $\alpha$, $\beta$, $\gamma$, and $\delta$ control the relative importance of each factor, with $\delta$ specifically tuning the preference for including branch points in the final stem path.
			
			\item \emph{Longest Path Computation:} To extract the main stem, we apply dynamic programming on a topologically sorted DAG to find the path with minimal total cost (equivalent to the "longest path" in a positive-cost formulation). Starting from the topmost node, we iteratively compute the minimum-cost path to each reachable node and track predecessors for path reconstruction.
		\end{itemize}

		\item Once the trunk path is extracted, we label nodes along this path as "stem" and others as "leaf." We then apply multiple refinement strategies:
		
		\begin{itemize}
			\item \emph{Immediate Neighborhood Analysis:} Branch-off nodes must have a minimum number of both stem and leaf neighbors to maintain their classification, otherwise, they are relabeled based on majority neighborhood type.
			\item \emph{Trunk Distance Analysis:} Branch-off nodes close to each other along the trunk path are compared by their reachable leaf sets, and if the overlap ratio exceeds a threshold, one is reclassified as a stem node to avoid duplicate detection of the same branch.
			\item \emph{Outlier Detection:} Isolated nodes without sufficient connectivity are flagged as outliers.
		\end{itemize}
		
		\item At each node of degree 3 we determine the next 5 nodes above and below. For each set of such 5 nodes we can calculate a linear regression line to approximate the trajectory of the leaf and the steam around the branching point (see Fig.~\ref{fig:leaf-angle-calculation}).
		
		\item Finally, the angle between the leaf and the stem is determined by converting the regression lines into 3D vectors \(\mathbf{v}_s\), \(\mathbf{v}_l\) and calculating \(\theta \;=\; \arccos\!\Bigl(\frac{\mathbf{v}_s \cdot \mathbf{v}_l}{\|\mathbf{v}_s\|\;\|\mathbf{v}_l\|}\Bigr)\)
		
	\end{enumerate}

	This method works well for plants that grow generally "upwards", but has limitations when leaves or parts of the plant grow in an arc, meaning that from stem to the tip of the leaf the z-coordinate is not strictly increasing. We have applied this metric thus only for young wheat plants of 14 days age after planting. At this growing stage (two-leaves) the method is still robust.
	
	
	\section{Data Analysis and Comparison with Manual Grading}\label{sec:data_analysis}
	In this section we showcase the overall utility of our photogrammetry system for phenotyping. We use the dataset described in Section~\ref{sub:data-collection} and had the plants manually rated by an expert with respect to their overall plant architecture. The expert rated each variety based on the UPOV scale~\cite{Richards2019} that ranges from  \(y\in\{1,\ldots,10\}\), describing whether the plants of that variety belong to a erectophile or planophile phenotypes. Here a score of 1 means fully erectophile and 10 means fully planophile. For this analysis, the expert's rating, which was given before the point clouds had been created, is considered as ground truth and we want to predict \(y\) from the features measured by the photogrammetry station as described above. Note that, while the expert has assigned a single rating to all plants of a single variety, the predictions are performed on each individual plant instead.
	
	We also point out that this analysis is for the purpose of proving the concept that the data collected by our system contains meaningful features that can be directly used to drive research questions (for example in a breeding program), rather than predicting manual measurements that have their own human bias and variance. In a production system, we anticipate the features extracted to \emph{augment} or even \emph{replace} manual measurements instead of predicting them. Futhermore, due to the relatively small size of the dataset and number of possible features we decided on two models: A multiple linear regression model and a \(k\)-nearest neighbor model. 
	
	\subsection{Rating via Multiple Linear Regression}
	Here we use a multiple linear regression model to fit training data to the assigned rating. We do this twice: Once for the point clouds that had been captured on day 14 after seeding and once for day 35 after seeding. We have split the full dataset (60 plants for each imaging day) into a training set of 48 plants (8 per variety) and a test set of 12 plants (2 per variety). The features considered are given in Table \ref{tab:features-considered}. 
	
	\begin{table*}
        \normalsize
		\centering

		\begin{tabular}{c p{0.6\linewidth}}
            \hline
            \textbf{Parameter} & \textbf{Description} \\
			\hline
			\(H_{\text{Max}}\)   & Plant height \\
			\hline
			\(R_{\text{Max}}\)   & Plant radius \\
			\hline
			\(V_{100}\)            & Volume of the entire plant's convex hull \\
			\hline
			\(V_{60}\), \(V_{40}\)   & Volume of the convex hull of the plant's top-60\% and top-40\% points, respectively \\
			\hline
			\(G\)                & Area of the alpha-shape of the ground cover projection (only for 35-days old plants) \\ 
			\hline
			\(\frac{H_{\text{Max}}}{V_k}\), \(\frac{R_{\text{Max}}}{V_k}\), \(\frac{V_100}{V_60}\), \(\frac{V_{100}}{V_{40}}\), \(\frac{V_{60}}{V_{40}}\) & Ratios of the above features (\(k=40,60,100\)) \\
			\hline
		\end{tabular}
		\caption{Features considered in prediction models.}
		\label{tab:features-considered}
	\end{table*}
	
	To select features, we perform an approach that starts with no features selected and consists of forward- and backward-steps (see for example \cite{james2023introduction}). In a forward step, each feature is temporarily added as predictor variable to the model and we select the feature that reduces the residual sum of squares on the training set the most. Then an immediate backward-step is performed in which we check whether with the newly added feature any of the predictor variables has a \(p\)-value that exceeds 0.05. If that is the case, these predictor variables are removed from the model and we iterate the procedure. Once no feature can be added to the model without being immediately removed again in the backward step the procedure terminates. The results of the such obtained model applied to the test set is given in Table \ref{tab:results-mlr}. Using this algorithm, we found for 14-day old point clouds that the features \(\frac{H_{\text{Max}}}{V_{60}}\), \(H_{\text{Max}}\), \(V_{100}\), and \(\frac{H_{\text{Max}}}{V_{40}}\) had been selected. For the point clouds of 35-days old plants the features \(G\), \(\frac{R_{\text{Max}}}{V_{100}}\), \(\frac{H_{\text{Max}}}{V_{100}}\), and \(\frac{H_{\text{Max}}}{R_{\text{Max}}}\) had been selected. 
	
	\begin{table}[h]
		\centering
        \normalsize
		\begin{tabular}{l l | l l l}
            \hline
			\textbf{Features Selected} & \textbf{\bm{$p$}-value} & \textbf{\bm{$R^2$}} & \textbf{MAE} & \textbf{Age (days)}\\
			\hline
			$\frac{H_{\text{Max}}}{V_{60}}$  & 0.008 & \multirow{4}{*}{0.72} & \multirow{4}{*}{1.73} & \multirow{4}{*}{14} \\
			\cline{1-2}
			$H_{\text{Max}}$  & \textless0.001 &  &  &\\
			\cline{1-2}
			$V_{100}$ & 0.005 & & &\\
			\cline{1-2}
			$\frac{H_{\text{Max}}}{V_{40}}$ & 0.033 & & &\\
			\hline \hline
			$G$  & <0.001 & \multirow{4}{*}{0.44} & \multirow{4}{*}{2.07} & \multirow{4}{*}{35} \\
			\cline{1-2}
			$\frac{R_{\text{Max}}}{V_{100}}$  & <0.001 &  &  &\\
			\cline{1-2}
			$\frac{H_{\text{Max}}}{V_{100}}$ & 0.001 & & &\\
			\cline{1-2}
			$\frac{H_{\text{Max}}}{R_{\text{Max}}}$ & 0.045 & & &\\
            \hline
		\end{tabular}
		\caption{Result of the multiple linear regression model reported on the held out test of the point clouds, for 14-days old plants (top part) and 35-days old plants (bottom part).}
		\label{tab:results-mlr}
	\end{table}

	\subsection{Rating via \(k\)-nearest neighbor model}
	In a second approach to predict the assigned plant architecture rating, we use a \(k\)-nearest neighbor model (\(k\)-NN) \cite{knn}. We split the data as before into a training and test set. Similar to the multiple linear regression model, we start investigating \(k\)-NN models with a single feature only, for which we optimize the hyperparameter \(k\) via 4-fold cross validation on the training set. To determine which feature to start the model with we use 4-fold cross-validation again. We then add features into the model, one by one, following the same methodology as long as the average mean-squared-error improves. After this feature-selection method and optimization of \(k\), we apply the model to the held-out test sets, respectively (14-day old plants and 35-day old plants). The selected features and performance on the test sets are presented in Table \ref{tab:results-knn}.
	
	\begin{table}[h]
		\centering
        \normalsize
		\begin{tabular}{l l l l l}
            \hline
			\textbf{Features Selected} & \(\bm{k}\) & \textbf{\bm{$R^2$}} & \textbf{MAE} & \textbf{Age (days)}\\
			\hline
			$\frac{H_{\text{Max}}}{V_{100}}, V_{40}$  & 5 & 0.54 & 1.48 & 14 \\
			\hline \hline
			$R_{\text{Max}}, \frac{H_{\text{Max}}}{V_{40}}, \frac{V_{60}}{V_{40}}, \frac{V_{100}}{V_{40}}$  & 2 & 0.81 & 1.00 & 35 \\
            \hline
		\end{tabular}
		\caption{Result of \(k\)-NN reported on the held out test of the point clouds, for 14-days old plants (top part) and 35-days old plants (bottom part).}
		\label{tab:results-knn}
	\end{table}

	Overall, we can see that the models, which are based on the data collected through our photogrammetry system, relate to the manually assigned ratings. However, there are still discrepancies: For the 14-day old plants the multiple linear regression model performed better with a mean absolute error (MAE) of 1.73 on the test set. For the 35-day old plants the \(k\)-NN model performed better with an MAE of 1.00. These mean errors are likely to reduce if the models can be trained on larger datasets, as well as when the ground truth ratings are being averaged over more than one manual rating, as there is an unknown bias and variance in ratings performed by humans.
	
	\section{Conclusion}\label{sec:conclusion}
	We have presented a low-cost photogrammetry station that uses off-the-shelf hardware and open software. We showed how a variety of phenotypic measurements can be extracted for wheat plants from the resulting 3D point clouds. We further demonstrated the system's capabilities by confirming that theses measurements, and other quantities derived from them, do indeed relate to the assessment of the plants as erectophile or planophile varieties. This underscores the immense potential of the system and the value it can generate for plant researchers and plant breeders. 
	
	Resources to reconstruct the system are available at our GitHub page \cite{githubrepo} containing a parts list, wiring diagrams, and all code used to produce the results described in this paper. The raw RGB-data and the reconstructed point clouds are available through the Federated Research Data Repository (FRDR) under the DOI: 10.20383/103.01255. 
    
   For future work, we are investigating the effects of different colored backdrops and lighting conditions, as well as a more cost-effective replacement for the commercial turntable. We also plan to develop new algorithms to extract additional plant features and to achieve a higher degree of robustness with the existing algorithms  presented here.  This will allow us to explore and analyze more complex plant architectures, as well as to tackle multiple plant images at the same time with out explicit segmentation and separation. Finally, a notable increase in applicability will be achieved by designing an in-field version of the system. 

    \section*{CRediT authorship contribution statement}
    \textbf{Joe Hrzich:} Methodology, Software, Investigation, Data Curation, Writing Original Draft, Visualization. \textbf{Michael A. Beck:} Methodology, Validation, Formal Analysis, Investigation, Writing -- Original Draft, Writing Review \& Editing, Supervision. \textbf{Christopher P. Bidinosti:} Conceptualization, Methodology, Resources, Writing Review \& Editing, Supervision, Project administration, Funding acquisition. \textbf{Christopher J. Henry:} Writing Review \& Editing, Project administration, Funding acquisition. \textbf{Kalhari Manawasinghe:} Conceptualization, Investigation, Resources. \textbf{Karen Tanino:} Writing Review \& Editing, Supervision..

    \section*{Declaration of competing interest}
    The authors declare that they have no known competing financial interests or personal relationships that could have appeared to influence the work reported in this paper. 

    \section*{Acknowledgements}
    This research was supported by Mitacs (Grant numbers IT25876, IT32715); NSERC Alliance, Saskatchewan Wheat Development Commission, Alberta Grains, Manitoba Crop Alliance (Grant number ALLRP 570351-2021).
    
	\bibliographystyle{unsrt}
	\bibliography{bibliography}
	
\end{document}